\documentclass[11pt]{article}

\usepackage{acl}

\usepackage{times}
\usepackage{latexsym}
\usepackage{amsmath}
\usepackage{amsfonts}
\usepackage{paralist}
\usepackage{booktabs}
\usepackage{subcaption} 
\usepackage{adjustbox}
\usepackage{array}
\usepackage{multirow}
\usepackage{fancyvrb}

\usepackage[T1]{fontenc}

\usepackage[utf8]{inputenc}

\usepackage{microtype}

\usepackage{inconsolata}

\usepackage{graphicx}
\usepackage{xcolor}
\newcommand{\best}[1]{\textcolor{red}{#1}}
\newcommand{\second}[1]{\textcolor{blue}{#1}}
\newcommand{\third}[1]{\textcolor{green}{#1}}

\newcommand{\CUT}[1]{}

\newcommand{\bz}{\mathbf{z}}
\newcommand{\bo}{\mathbf{o}}
\newcommand{\bw}{\mathbf{w}}

\newcommand{\bx}{\mathbf{x}}
\newcommand{\bH}{\mathbf{H}}

\newcommand{\real}{\mathbb{R}}

\title{Faithfulness Evaluation for Decoder-only LLM Attributions with Controlled Retained Information}

\author{Xin Huang \\
  City University of Hong Kong\\
  Department of Computer Science\\
 \
  \texttt{xhuang899-c@my.cityu.edu.hk} \\\And
  Antoni B. Chan \\
  City University of Hong Kong\\
  Department of Computer Science\\
 \
  \texttt{abchan@cityu.edu.hk} \\}

\begin{document}
\maketitle 
\begin{abstract}
Large Language Models (LLMs) are increasingly evaluated with input attribution methods, yet comparing such explanations remains challenging. Existing soft-perturbation faithfulness metrics, such as Soft-NC and Soft-NS, can conflate attribution quality with the number of words retained during perturbation: attribution methods with larger average scores may keep more words and therefore obtain inflated scores. To address this issue, we propose $\pi$-Soft-NC and $\pi$-Soft-NS, an evaluation framework that compares attribution methods under the same expected retaining probability, thus controlling the number of retained words. We further introduce Grad-ELLM, a gradient-based attribution method tailored to autoregressive decoder-only LLMs, which combines gradient-derived channel importance with attention-derived token importance at each decoding step. Experiments on classification and open-generation tasks with Llama and Mistral show that Grad-ELLM achieves strong comprehensiveness-oriented faithfulness under $\pi$-Soft-NC, while there is no dominant method under $\pi$-Soft-NS.
Our evaluation metric serves as a rigorous framework to compare XAI methods for LLMs, which will support progress in the field.

\end{abstract}

\section{Introduction}

Large language models such as Llama have demonstrated their remarkable capabilities across multiple downstream tasks, e.g., text generation and sentiment classification \cite{grattafiori2024llama3herdmodels}. However, as these models grow in size and complexity, their ``black box'' property has raised significant concerns regarding transparency, trustworthiness, and robustness. The lack of interpretability in LLMs hinders their trustworthiness and thus adoption in high-stakes domains, e.g., healthcare \cite{esteva2019guide} and finance \cite{veale2017fairer}, where users need not only accurate predictions but also faithful evidence about the model's decision process. Input attribution methods address this need by assigning importance scores to input tokens \citep{sundararajan2017axiomatic,lei-etal-2016-rationalizing}, where the reliability of the attributions is measured using faithfulness metrics. 

Faithfulness is commonly evaluated by perturbing inputs according to attribution scores and measuring the resulting output change \citep{DBLP:conf/bmvc/PetsiukDS18,deyoung-etal-2020-eraser}. Hard perturbation metrics, such as deletion, insertion, comprehensiveness, and sufficiency, are intuitive but can create out-of-distribution inputs when tokens are entirely removed or retained \citep{ancona2018towards,bastings-filippova-2020-elephant,yin2022sensitivity}. Normalized Soft Comprehensiveness (Soft-NC) and Normalized Soft Sufficiency (Soft-NS) \citep{zhao2024reagent} alleviate this issue by evaluating continuous attribution distributions through soft perturbations.


However, we observe that Soft-NC/NS can yield unfair comparisons when different attribution methods produce score distributions with different means. Since the attribution score directly determines the probability of retaining or removing each token, a method with larger average scores can keep more input information during perturbation. As a result, its faithfulness score may improve simply because the perturbed input contains more information, rather than because the attribution is more faithful. This issue is especially important for decoder-only LLM explanations, where attribution methods can produce highly different score distributions across generation steps.

To address this issue, we propose $\pi$-Soft-NC and $\pi$-Soft-NS, which evaluate attribution methods under a fixed target retaining probability $\pi$. Instead of comparing raw attribution scores directly, we transform each attribution distribution so that different methods retain the same expected amount of information during soft perturbation. This allows faithfulness comparisons to focus on the quality and calibration of the attribution distribution, rather than on differences in perturbation strength. Moreover, sweeping $\pi$ reveals how attribution evidence is organized: high $\pi$-Soft-NS at small retaining probabilities indicates compact sufficient evidence, while strong $\pi$-Soft-NC across retaining probabilities indicates broader comprehensiveness coverage.

In addition to the evaluation framework, we propose Grad-ELLM, a gradient-based attribution method tailored to autoregressive decoder-only LLMs. Grad-ELLM adapts the gradient-attention aggregation idea of Grad-ECLIP \citep{zhao2024gradient} to sequential text generation. Unlike CLIP \citep{radford2021learning}, which uses a dual-encoder architecture and a global image-text alignment objective, decoder-only LLMs generate tokens sequentially. Grad-ELLM therefore computes attribution with respect to the next-token logit at each decoding step by combining gradient-derived channel importance with attention-derived token importance, and then aggregates attribution information across generation steps.



In summary, our contributions are four-fold:
1) We identify a bias in existing Soft-NC/NS metrics: attribution methods with different score means can be evaluated under different expected amounts of retained information.
2) We propose $\pi$-Soft-NC and $\pi$-Soft-NS,  faithfulness metrics that control the retained information by comparing attribution methods under the same target retaining probability.
3) We introduce Grad-ELLM, a gradient-based attribution method for autoregressive decoder-only LLMs that combines gradient-derived channel importance with attention-derived token importance.
4) We conduct a comprehensive comparison of representative LLM attribution methods under multiple faithfulness protocols, showing that $\pi$-Soft-NC/NS reveal different evidence structures, including compact sufficient evidence, broad comprehensiveness-oriented evidence, and dense but unselective attribution distributions.

\section{Related Work}

XAI for LLMs includes both local and global explanations \citep{zhao2024explainability,luo2024understanding}; 
we focus on local input attribution and its faithfulness evaluation for decoder-only LLMs.

\subsection{General Input Attribution Methods}

Input attribution methods seek to quantify the contribution of input features (e.g., tokens) to model outputs.
Early methods were model-agnostic and originated from computer vision or general ML contexts, but have been adapted to NLP. Gradient-based methods, including Saliency, Integrated Gradients, DeepLIFT, and Input$\times$Gradient \citep{DBLP:journals/corr/SimonyanVZ13,sundararajan2017axiomatic,shrikumar2017learning}, propagate output signals back to input features and are computationally efficient. However, applying them to decoder-only LLMs is challenging because attribution must be computed across autoregressive generation steps, where noise can accumulate and score distributions can vary substantially across methods \citep{zhao2024reagent,fayyaz2024evaluatinghumanalignmentmodel}.
%
%
Perturbation-based methods, such as LIME \citep{ribeiro2016should}, estimate importance by altering inputs and observing output changes, but text perturbations can break semantic coherence and introduce out-of-distribution artifacts.

Our proposed Grad-ELLM extends gradient-based methods to decoder-only LLMs, incorporating attention mechanisms to handle autoregressive generation without additional forward passes, thus maintaining efficiency while enhancing faithfulness in sequential outputs.

\subsection{Attributions for LLMs/Transformers}

%

Recent LLM attribution methods address autoregressive generation and long-context settings. Value Zeroing \citep{mohebbi-etal-2023-quantifying} directly measures the effect of zeroing token representations, but requires many forward passes, making it computationally heavy. ReAGent \citep{zhao2024reagent} uses an auxiliary RoBERTa model \citep{liu2019robertarobustlyoptimizedbert} to generate semantically preserving perturbations, improving perturbation quality at the cost of additional computation. These methods differ not only in computational cost, but also in the scale and density of their attribution scores, which can affect faithfulness evaluation.

%
%

Attention-based explanations are efficient because they reuse internal transformer weights \citep{vaswani2017attention,DBLP:journals/corr/BahdanauCB14}, but raw attention is not always faithful to model decisions \citep{Jain2019AttentionIN,serrano-smith-2019-attention}. Gradient-attention variants such as Grad-SAM \citep{barkan2021grad} improve over raw attention by incorporating gradient signals, but decoder-only generation still requires step-wise attribution.

%

%

For VLMs, Grad-ECLIP \citep{zhao2024gradient} explains CLIP \citep{radford2021learning} by combining gradient-derived channel weights with softened attention maps. It visualizes image-text alignments in a dual-encoder vision-language model, but does not directly address the sequential decoding process required for autoregressive LLMs. Grad-ELLM extends this idea from dual-encoder image-text alignment to autoregressive next-token prediction, where attribution must be computed and evaluated across generation steps.



\subsection{Evaluation of Attribution Faithfulness}

Faithfulness is commonly assessed through perturbation-based metrics, including insertion/deletion curves \citep{DBLP:conf/bmvc/PetsiukDS18}, comprehensiveness/sufficiency \citep{deyoung-etal-2020-eraser}, and Area Over the Perturbation Curve (AOPC) \citep{samek2016evaluating}.
These metrics test whether perturbing important input features causes large changes in model behavior. 
\citet{zhao2024reagent} propose Soft-NC and Soft-NS to handle continuous attributions. However, these metrics may unfairly penalize methods with different information loss profiles during perturbation (see \S\ref{sec:pisoft}).


Other evaluation metrics, such as FaithScore \citep{jing2024faithscore} and Ragas \citep{es2024ragas}, target factuality or RAG grounding rather than token-level attribution faithfulness. 
Our $\pi$-Soft-NC/NS metrics build on these faithfulness evaluations by introducing a normalization for expected information loss, ensuring fairer comparisons across attribution methods that generate score distributions with different characteristics, particularly in autoregressive LLM contexts. This control also makes the resulting $\pi$-curves more interpretable: they reveal whether an attribution method captures compact sufficient evidence or broad comprehensiveness-oriented evidence.

\section{Methodology}
We first propose our $\pi$-Soft-NC/NS faithfulness metrics.
We then give a brief overview of decoder-only LLMs, and then propose our corresponding Grad-ELLM attribution method.

\subsection{$\pi$-Soft-NC and $\pi$-Soft-NS}
\label{sec:pisoft}
We propose a modification of Soft-NC/NS, denoted as $\pi$-Soft-NC/NS, to obtain a fairer and more comprehensive faithfulness metric.

{\bf Soft-NC/NS.}
\citet{zhao2024reagent} proposed  
Soft-NS
and Soft-NC
to evaluate the faithfulness of the entire attribution distribution. They argue that  traditional sufficiency and comprehensiveness metrics 
are hindered by an out-of-distribution issue \citep{ancona2018towards, bastings-filippova-2020-elephant, yin2022sensitivity}. Specifically, hard perturbations---which involve entirely removing or retaining tokens---generate discretely corrupted versions of the original input that may fall outside the model's training distribution. As a result, the model's predictions on these altered inputs are unlikely to align with the reasoning processes used for the original full sentences, potentially leading to misleading insights into the model's mechanisms.

To mitigate this issue, rather than create hard perturbations by thresholding or ranking the attribution scores (as in insertion/deletion metrics), soft perturbations are generated by sampling perturbations according to the attribution scores.
To create a soft perturbation of the input text, an independent Bernoulli distribution is applied to randomly mask each input embedding token vector $\bx_i$,
according to its attribution score $s_i$, 
%
\begin{align}
     \bx'_i = e_i\bx_i  , \ \ e_i \sim \mathrm{Bernoulli}(s_i).
\end{align}
Here we assume $s_i\in[0,1]$, which can be viewed as introducing information loss (perturbation) at the embedding level: if $s_i=0$, the embedding vector $\bx_i$ is zeroed out (complete masking); if $s_i=1$, the embedding is kept  (no masking).
We denote the attribution scores for the whole text as $\mathbf{s}=\{s_1,\cdots,s_m\}$.

Denote $X'=\{x'_1,\cdots,x'_m\}$ as the soft-perturbation text based on the attribution scores for the next generated token $y_t$.
Let $\textbf{P}_{X,t}=[p_{1,t}, \cdots, p_{v,t} ]$ be the probability distribution of $y_t$ over the whole vocabulary (of size $v$) when the input of the model is the original text $X$, and similarly 
$\textbf{P}_{X',t}=[p'_{1,t}, \cdots, p'_{v,t} ]$ for when the input is the soft-perturbed text $X'$. The effect of the perturbation on the output distribution is measured using the Hellinger distance:
\begin{align}
    \Delta \textbf{P}_{X',t} & = H( \textbf{P}_{X,t}, \textbf{P}_{X',t}) 
     \\
     &= \tfrac{1}{\sqrt{2}}\Big[\sum_{i=1}^{v}(\sqrt{p_{i,t}} - \sqrt{p'_{i,t}})^2\Big]^{1/2}   ,
\end{align}

Finally, Soft-NS is defined as the relative effect of keeping important tokens:
\begin{equation}
    \operatorname{Soft-NS}(X,x_t,\mathbf{s}) = \tfrac{\max(0, \Delta \textbf{P}_{0,t} - \Delta \textbf{P}_{X',t})}{\Delta \textbf{P}_{0,t} },
\end{equation}
where $\Delta \textbf{P}_{X',t}$ is the effect of the soft-perturbation $X'$, and $\Delta \textbf{P}_{0,t}$ is the effect of the zero baseline (all zero inputs).
Likewise, 
Soft-NC
is defined as the relative effect of removing important tokens,
\begin{equation}
    \operatorname{Soft-NC}(X,x_t,\mathbf{s}) = \tfrac{\Delta \textbf{P}_{\bar{X}',t}}{\Delta \textbf{P}_{0,t} },
\end{equation}
where $\bar{X}'$ is the soft-perturbation that \emph{removes} tokens, i.e., with masks $e_i \sim \mathrm{Bernoulli}(1-s_i)$, and $\Delta \textbf{P}_{\bar{X}',t}$ is its corresponding effect. Since Soft-NC is normalized by the zero-baseline distance $\Delta P_{0,t}$, its value can exceed 1 when the perturbed input induces a larger distributional shift than the zero baseline.

{\bf Bias of Soft-NC/NS.} 
Soft-NC/NS has a problem that makes it potentially unfair to compare the performance of different attribution methods. 
Different attribution methods may yield heatmaps with different mean values, and thus  the overall probability of keeping any word (or the expected number of kept words) after soft perturbation may also differ. 
In particular, given the text attribution scores $\mathbf{s}$, the probability of retaining any word is:
    \begin{align}
    \hat{\pi} = \tfrac{1}{m}\sum\nolimits_{i=1}^{m}\mathbb{E}[e_i]
    = \tfrac{1}{m}\sum\nolimits_{i=1}^{m} s_i,
    \end{align}
and the expected number of kept words is $m\hat{\pi}$.  It will be unfair to compare the Soft-NC/NS values for two attribution methods $A$ and $B$ with different retaining probabilities, $\hat{\pi}_a$ and $\hat{\pi}_b$, as the expected amount of retained information is different.  That is, the difference in Soft-NC/NS scores may be due to different amounts of attributed words, rather than better attribution quality.

{\bf $\pi$-Soft-NC/NS.} 
To address the above problem, we propose 
$\pi$-Soft-NC/NS, 
which first transforms the attribution scores so that the calculated retaining probability $\hat{\pi}$ matches a given target probability $\pi\in[0,1]$.
Assuming scores $s_i\in[0,1)$, we adopt the alpha transformation, $\tilde{s}_i = s_i^\alpha$, and set $\alpha$ such that the transformed scores match the target retaining probability, $\frac{1}{m}\sum_{i=1}^m s_i^\alpha = \pi$. Note that target $\pi=1$ is obtained by setting $\alpha=0$, and $\pi=0$ is obtained with $\alpha \rightarrow \infty$. We then define: 
    \begin{align}
    \nonumber
         &\pi\operatorname{-Soft-NC}(X,x_t,\mathbf{s},\pi) = 
         \operatorname{Soft-NC}(X,x_t,\mathbf{s}^\alpha), \\
    \nonumber
         &\pi\operatorname{-Soft-NS}(X,x_t,\mathbf{s},\pi) = 
         \operatorname{Soft-NS}(X,x_t,\mathbf{s}^\alpha), \\
    &\quad \text{where } \tfrac{1}{m}\sum\nolimits_{i=1}^m s_i^\alpha = \pi.
    \end{align}
In practice, $\alpha$ is solved numerically, e.g., using bisection search. For $\pi$-Soft-NS, $\pi$ directly corresponds to the expected fraction of retained input information. For $\pi$-Soft-NC, the same transformed attribution scores are first calibrated to target $\pi$ and then complemented to remove high-attribution information -- larger $\pi$ indicates that a larger expected mass of high-attribution information is removed.

\begin{figure*}[t]

\includegraphics[width=\textwidth]{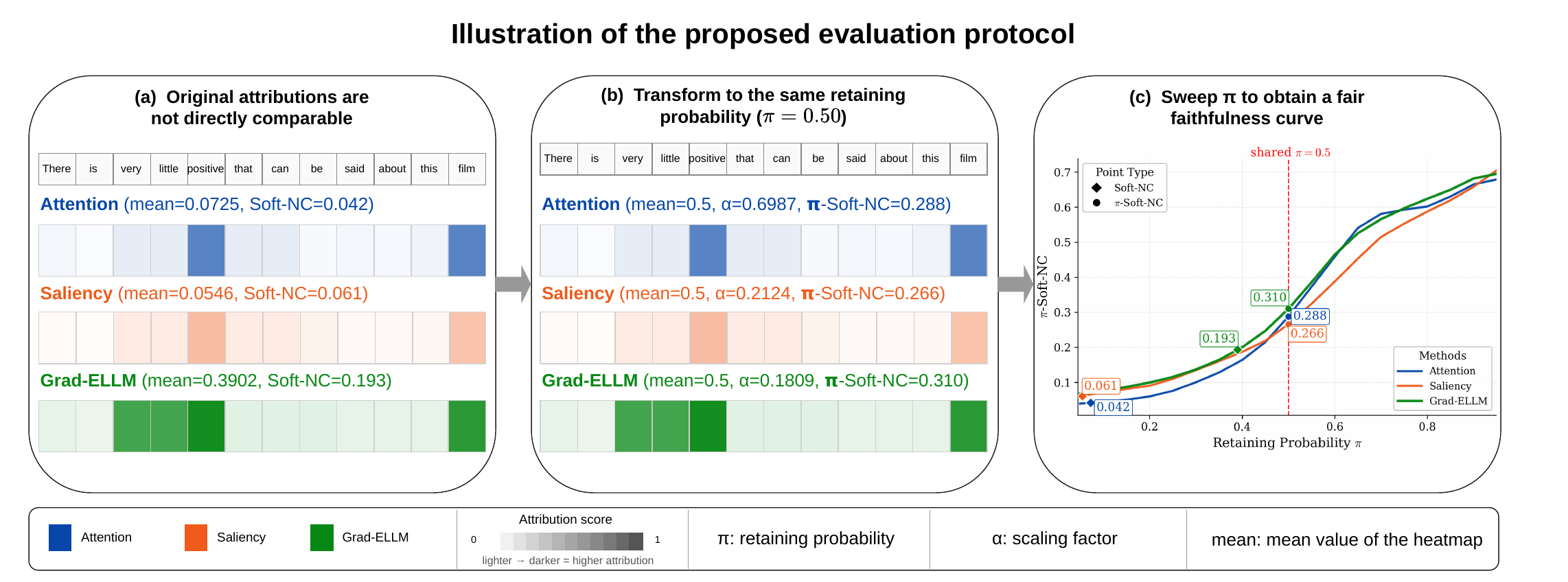}
\caption {Illustration of the proposed $\pi$-Soft evaluation protocol. (a) Different attribution methods can produce score distributions with different means, leading to different expected retaining probabilities and therefore incomparable original Soft-NC values.
(b) For a target retaining probability $\pi$, each attribution distribution is transformed with its own scaling parameter $\alpha$ so that all methods share the same expected retaining probability; the resulting $\pi$-Soft-NC values are therefore comparable.
(c) Sweeping $\pi$ yields the full $\pi$-Soft-NC curve. Original Soft-NC values correspond to the special case $\alpha=1$ and appear as different points on the curve, showing why they are not directly comparable across methods.}
  \label{fig:illustration_evaluation_protocol}
\end{figure*}

Using the same target $\pi$, two attribution methods can be fairly compared using $\pi$-Soft-NC/NS, where both soft-perturbed text will now have the same expected number of retained words. 
Because the soft-perturbation is based on attribution scores, attribution methods that perform well according to $\pi$-Soft-NC/NS will produce scores that are well-calibrated probabilities of importance. 
Furthermore, a comprehensive evaluation is obtained by sweeping the value of $\pi \in [0,1]$ and plotting $\pi$-Soft-NC/NS vs. $\pi$ curves, and then summarizing using AUC. The illustration of the proposed evaluation protocol is shown in Fig.~\ref{fig:illustration_evaluation_protocol}.

{\bf Evaluation Protocols.}
The attention mechanism in Transformers dynamically routes information in each layer based on the embeddings of the previous layers. Thus, we propose two evaluation protocols to consider different aspects of evaluation.

\begin{compactitem}
    \item \emph{Dynamic-Routing}: the transformer attention is dynamically computed based on the perturbed inputs. 
    The measured output change reflects both the causal contribution of token content and the model's dynamic re-routing after perturbation.
    \item \emph{Fixed-Routing}: the transformer attention is fixed based on the non-perturbed input (i.e., the original forward pass). Here changes in the output distribution are mainly caused by changes to the value/content information carried by the perturbed tokens, rather than by attention reallocation.
\end{compactitem}
In summary, the dynamic-routing protocol evaluates how the attributions identify words that \emph{could} be useful for the model under dynamic conditions, while the fixed-routing protocol is a local evaluation of the attributed words under the fixed context of the original input.

\subsection{Preliminary on LLMs}
Decoder-only LLMs, such as Llama \cite{grattafiori2024llama3herdmodels} and Mistral \cite{jiang2023mistral7b}, generate text autoregressively: given an input sequence of tokens $X = \{x_1,x_2,\dots, x_m\}$, the model predicts output tokens $Y = \{y_1,y_2,\dots,y_n\}$ sequentially, where each $y_t$ is conditioned on all the inputs and the previous outputs $\{x_1,\dots,x_m,y_1,\dots,y_{t-1}\}$. That is, at step $t$, the model computes the logit $l_t$ for $y_t$ based on the prefix $\{x_1, \dots, x_m, y_1, \dots, y_{t-1}\}$. 

The Transformer decoder consists of $N$ layers, each with self-attention and feed-forward networks. In self-attention, for a query $\mathbf{q}$, keys $\mathbf{k}_i$, and values $\mathbf{v}_i$, the attention weights are:
\begin{equation}
\label{spatial weight}
    \lambda_i = \operatorname{softmax}(\tfrac{\mathbf{q} \cdot \mathbf{k}_i^\top}{\sqrt{d}}) ,
\end{equation}
and the output is $\sum_i \lambda_i \mathbf{v}_i$, where $d$ is the embedding dimension. The attention weight captures token dependencies, but does not necessarily reflect actual contributions to the output \cite{Jain2019AttentionIN}, which necessitates explanation methods. 

\begin{figure*}[t]
  \includegraphics[width=\textwidth]{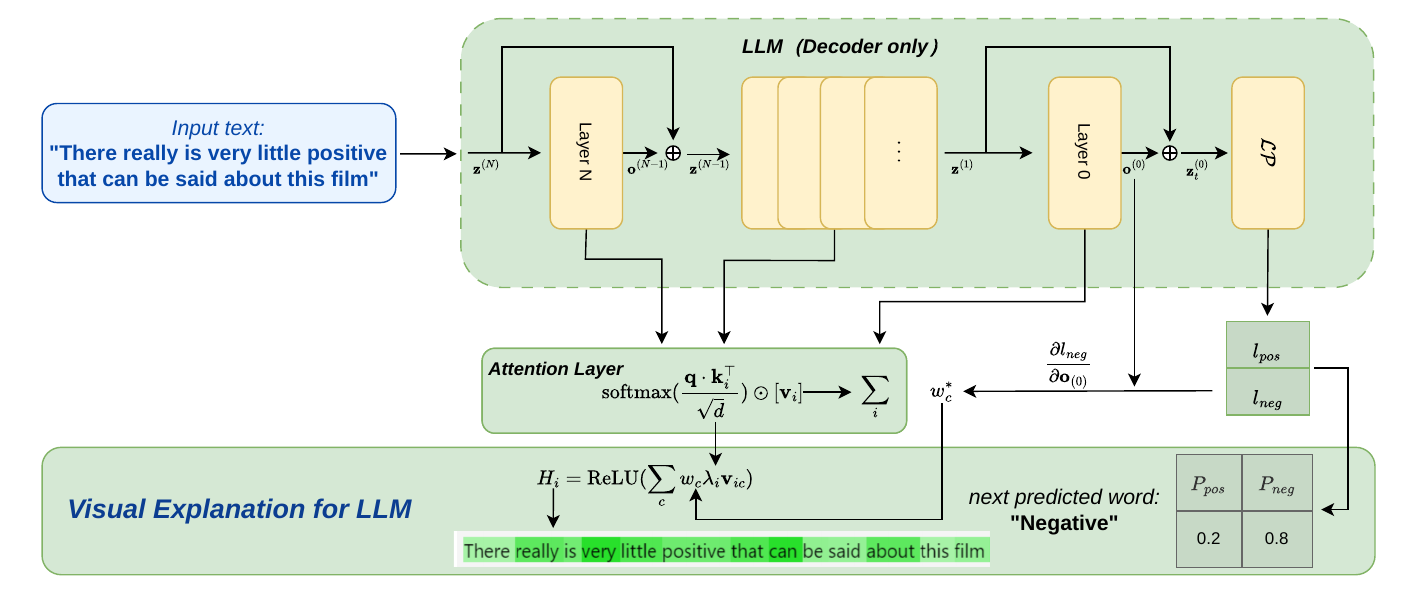}
  \caption {Framework of the proposed Grad-ELLM. An attribution map is generated by aggregating the values $\mathbf{v}$ as feature map in the attention layer with token weight $\lambda_i$ and channel weight $w_c$. This example shows the attribution map when generating the word ``Negative''.}
  \label{framework}
\end{figure*}

We denote $\bz^{(k)}\in\real^{(m+t) \times d}$ as the input to layer $k$ (output of layer $k+1$), with $\bz^{(N)}$ as token embeddings (of inputs $X$) and $\bz^{(0)}$ as the final hidden states (see Fig.~\ref{framework}).
The logit $l_t$ for the next token $y_t$ is derived from the $t$-th generated token's hidden state and $\bz^{(0)}_t\in\real^d$ is the corresponding row in $\bz^{(0)}$. We denote $\bo^{(k)}\in\real^{(m+t) \times d}$ as the output of the self-attention block in layer $k$, and thus the output of the transformer block $k$ is $\bz^{(k-1)} = \bo^{(k-1)} + \bz^{(k)}$.

\subsection{Gradient-based Explanations for LLMs (Grad-ELLM)}
\label{sec:grad-ellm}

In addition to the proposed evaluation metrics, we introduce Grad-ELLM as a gradient-based attribution method for autoregressive decoder-only LLMs. Grad-ELLM explains the logit $l_t$ of the next generated token $y_t$ by combining two sources of information in the self-attention computation: gradient-derived channel importance and attention-derived token importance.

For a transformer layer $k$, let $\bo_t^{(k)}$ be the attention output at the last query position when predicting $y_t$. The attention output can be written as
\begin{equation}
    \bo_t^{(k)} = \sum_i \lambda_i^{(k)} \mathbf{v}_i^{(k)},
\end{equation}
where $\mathbf{v}_i^{(k)}$ is the value vector of the $i$-th input token, and $\lambda_i^{(k)}$ is its token weight derived from the query-key similarity. We approximate the contribution of each channel by the gradient of the target logit with respect to the attention output:
\begin{equation}
    w_c^{(k)} = \frac{\partial l_t}{\partial \bo_t^{(k)}[c]}.
\end{equation}
The attribution assigned to the $i$-th token at layer $k$ is then
\begin{equation}
    H_i^{(k)} =
    \operatorname{ReLU}
    \left(
    \sum_c w_c^{(k)} \lambda_i^{(k)} \mathbf{v}_{ic}^{(k)}
    \right).
    \label{eq:grad-ellm-main}
\end{equation}
We aggregate $H_i^{(k)}$ over the selected decoder layers and normalize the resulting token scores to obtain the final attribution map. Following Grad-ECLIP \citep{zhao2024gradient}, we use normalized query-key similarities as loosened token weights, i.e., $\lambda_i^{(k)} \approx \Phi(\mathbf{q}_t^{(k)}\mathbf{k}_i^{(k)\top})$, where $\Phi$ denotes min--max normalization over input tokens. This avoids overly peaky softmax weights and yields a continuous attribution distribution over the input. The complete derivation is provided in Appendix~\ref{app:grad-ellm-derivation}.

\section{Experiments}
We evaluate both the proposed $\pi$-Soft-NC/NS metrics and Grad-ELLM. We first compare Grad-ELLM with representative attribution baselines under $\pi$-Soft-NC/NS, and then analyze complementary ranking-based behavior using insertion/deletion metrics in the Appendix~\ref{app:insertion/deletion}. 
%

\subsection{Setup}
We use Llama-3.1-8B-Instruct and Mistral-7B-Instruct-v0.3 from Hugging Face Transformers \citep{wolf2020transformers}. Evaluation on a larger Llama-3.1-70B-Instruct is reported in Appendix~\ref{app:larger_model}. See 
Appendix \ref{app:implementation} for implementation details.

We compare with representative baseline XAI methods from each category: 1) attention map-based methods, including raw attention in the last decoder layer \cite{DBLP:journals/corr/BahdanauCB14}; 
2) classical gradient-based methods, such as Saliency (vanilla gradients) \cite{DBLP:journals/corr/SimonyanVZ13}, Input$\times$Gradient \cite{DBLP:journals/corr/SimonyanVZ13}, Integrated Gradients (IG) \cite{sundararajan2017axiomatic}, and DeepLIFT \cite{shrikumar2017learning};
3) perturbation-based Value Zeroing \cite{mohebbi-etal-2023-quantifying}; 
4) a random baseline (Random), which assigns uniform random scores to tokens as a control.
5) Recursive Attribution Generator (ReAGent) \cite{zhao2024reagent}, a model-agnostic feature attribution method that exploits an external RoBERTa model.

\subsection{Datasets}
We evaluate on three classification datasets and two open-generation datasets following \citet{zhao2024reagent}. The classification datasets include IMDb \citep{maas2011learning}, SST2 \citep{socher2013recursive}, and BoolQ from ERASER \citep{deyoung-etal-2020-eraser}. The open-generation datasets include TellMeWhy \citep{lal-etal-2021-tellmewhy} and WikiBio \citep{manakul2023selfcheckgpt}. Detailed dataset descriptions and prompts are provided in Appendix~\ref{app:dataset} and~\ref{sec:appendix}.

\subsection{Faithfulness Metrics}

We mainly use our proposed $\pi$-Soft-NC and $\pi$-Soft-NS with \emph{dynamic-routing} protocol to measure the faithfulness of the full importance distribution fairly. 
We compute curves using $\pi \in \{0.05, 0.10, \dots, 0.95\}$. For each input and each retaining probability $\pi$, $\pi$-Soft-NC/NS are estimated by averaging over 15 Monte Carlo samples of the stochastic soft perturbation.

%
Following \citet{zhao2024reagent}, for classification tasks with explicit label words, we evaluate the \textbf{token-level faithfulness} of the target label word. For open-generation tasks, we evaluate sequence-level faithfulness by computing $\pi$-Soft-NC/NS every five generated tokens from the sequential prediction and averaging the resulting scores across evaluation steps.

Due to space limiations, evaluation under the \emph{fixed-routing} protocol is presented in Appendix~\ref{app:fixed-routing}.

\begin{table*}[htb]
\centering
\tiny
\setlength{\tabcolsep}{3.0pt}
\renewcommand{\arraystretch}{1.05}

\begin{subtable}[t]{0.49\textwidth}
\centering
\caption{Llama: AUC of $\pi$-Soft-NC/NS ($\uparrow$)}
\label{tab:llama-soft}
\begin{tabular}{lccccccccc}
\toprule
 & \multicolumn{9}{c}{Methods} \\
\cmidrule(lr){2-10}
Dataset & {Attn} & {DL} & {I$\times$G} & {IG} & {Rnd} & {Sal} & {VZ} & {RG} & {Ours} \\
\midrule
\multicolumn{10}{l}{\textbf{AUC $\pi$-Soft-NS $\uparrow$}}\\
IMDb      & 0.603 & 0.574 & 0.570 & 0.585 & 0.551 & 0.568 & 0.592 & 0.559 & 0.590 \\
SST2      & 0.592 & 0.588 & 0.588 & 0.577 & 0.549 & 0.577 & 0.594 & 0.562 & 0.586 \\
BoolQ     & 0.337 & 0.382 & 0.381 & 0.357 & 0.291 & 0.384 & 0.342 & 0.310 & 0.345 \\
TellMeWhy & 0.473 & 0.506 & 0.504 & 0.526 & 0.392 & 0.500 & 0.474 & 0.463 & 0.453 \\
WikiBio   & 0.458 & 0.476 & 0.475 & 0.476 & 0.352 & 0.472 & 0.471 & 0.401 & 0.459 \\
\textbf{Avg} & 0.493 & \best{0.505} & \third{0.504} & \second{0.504} & 0.427 & 0.500 & 0.495 & 0.459 & 0.487 \\

\midrule
\multicolumn{10}{l}{\textbf{AUC $\pi$-Soft-NC $\uparrow$}}\\
IMDb      & 0.433 & 0.404 & 0.402 & 0.410 & 0.412 & 0.399 & 0.420 & 0.405 & 0.437 \\
SST2      & 0.296 & 0.306 & 0.303 & 0.304 & 0.312 & 0.290 & 0.304 & 0.305 & 0.315 \\
BoolQ     & 1.574 & 2.228 & 2.195 & 2.153 & 1.888 & 2.221 & 1.557 & 1.901 & 2.398 \\
TellMeWhy & 0.754 & 0.756 & 0.749 & 0.693 & 0.622 & 0.749 & 0.736 & 0.542 & 0.649 \\
WikiBio   & 0.785 & 0.743 & 0.740 & 0.762 & 0.663 & 0.732 & 0.812 & 0.652 & 0.797 \\
\textbf{Avg} & 0.768 & \second{0.887} & \third{0.878} & 0.864 & 0.779 & 0.878 & 0.766 & 0.761 & \best{0.919} \\
\bottomrule
\end{tabular}

\vspace{2pt}
\captionsetup{font=footnotesize}
\end{subtable}
\hfill
\begin{subtable}[t]{0.49\textwidth}
\centering
\caption{Mistral: AUC of $\pi$-Soft-NC/NS ($\uparrow$)}
\label{tab:mistral-soft}
\begin{tabular}{lccccccccc}
\toprule
& \multicolumn{9}{c}{Methods} \\
\cmidrule(lr){2-10}
Dataset & {Attn} & {DL} & {I$\times$G} & {IG} & {Rnd} & {Sal} & {VZ} & {RG} & {Ours} \\
\midrule
\multicolumn{10}{l}{\textbf{AUC $\pi$-Soft-NS $\uparrow$}}\\
IMDb      & 0.495 & 0.482 & 0.484 & 0.463 & 0.401 & 0.479 & 0.526 & 0.425 & 0.481 \\
SST2      & 0.453 & 0.437 & 0.439 & 0.451 & 0.317 & 0.443 & 0.474 & 0.324 & 0.414 \\
BoolQ     & 0.360 & 0.356 & 0.357 & 0.352 & 0.312 & 0.356 & 0.367 & 0.407 & 0.367 \\
TellMeWhy & 0.479 & 0.483 & 0.483 & 0.485 & 0.384 & 0.486 & 0.491 & 0.463 & 0.465 \\
WikiBio   & 0.520 & 0.540 & 0.570 & 0.537 & 0.412 & 0.573 & 0.456 & 0.476 & 0.541 \\
\textbf{Avg} & 0.461 & 0.460 & \second{0.467} & 0.458 & 0.365 & \best{0.467} & \third{0.463} & 0.419 & 0.454 \\

\midrule
\multicolumn{10}{l}{\textbf{AUC $\pi$-Soft-NC $\uparrow$}}\\
IMDb      & 0.463 & 0.497 & 0.496 & 0.442 & 0.420 & 0.489 & 0.400 & 0.403 & 0.436 \\
SST2      & 0.336 & 0.384 & 0.384 & 0.435 & 0.466 & 0.374 & 0.314 & 0.458 & 0.431 \\
BoolQ     & 0.471 & 0.490 & 0.491 & 0.474 & 0.403 & 0.498 & 0.468 & 0.475 & 0.497 \\
TellMeWhy & 0.632 & 0.612 & 0.612 & 0.564 & 0.592 & 0.608 & 0.632 & 0.570 & 0.593 \\
WikiBio   & 0.633 & 0.600 & 0.482 & 0.578 & 0.615 & 0.481 & 0.765 & 0.604 & 0.645 \\
\textbf{Avg} & 0.507 & \second{0.517} & 0.493 & 0.499 & 0.499 & 0.490 & \third{0.516} & 0.502 & \best{0.520} \\
\bottomrule
\end{tabular}
\end{subtable}

\caption{Faithfulness evaluation with AUC of $\pi$-Soft-NC/NS curves for (a) Llama and (b) Mistral. \textcolor{red}{Best} / \textcolor{blue}{2nd} / \textcolor{green}{3rd} are highlighted on the \textbf{Avg} row. Abbr.: Attn = Attention, DL = DeepLIFT, I$\times$G = Input$\times$Gradients, IG = Integrated Gradients, Rnd = Random, Sal = Saliency, VZ = Value Zeroing, RG = ReAGent.}
\label{tab:soft-faithfulness-sidebyside}
\end{table*}

\CUT{\subsection{Implementation Details}
We use Llama-3.1-8B-Instruct and Mistral-7B-Instruct-v0.3 from Hugging Face Transformers \citep{wolf2020transformers}. A larger Llama-3.1-70B-Instruct evaluation is reported in the Appendix~\ref{app:larger_model}. Baselines are implemented with Inseq \citep{sarti-etal-2023-inseq}. We set \texttt{top\_p=1} for deterministic generation across attribution methods. Experiments with 7B/8B models are run on a single 48GB NVIDIA RTX6000 Ada GPU; the 70B model is evaluated on a single 80GB A100 with 4-bit quantization. Results are averaged over 15 random seeds. Additional implementation details are provided in Appendix~\ref{app:implementation}.
}

\subsection{Quantitative Results}

\begin{figure*}[htb]
  \centering
  \vspace{-0.8em}

  \begin{minipage}[t]{0.485\textwidth}
    \centering
    \includegraphics[width=\linewidth]{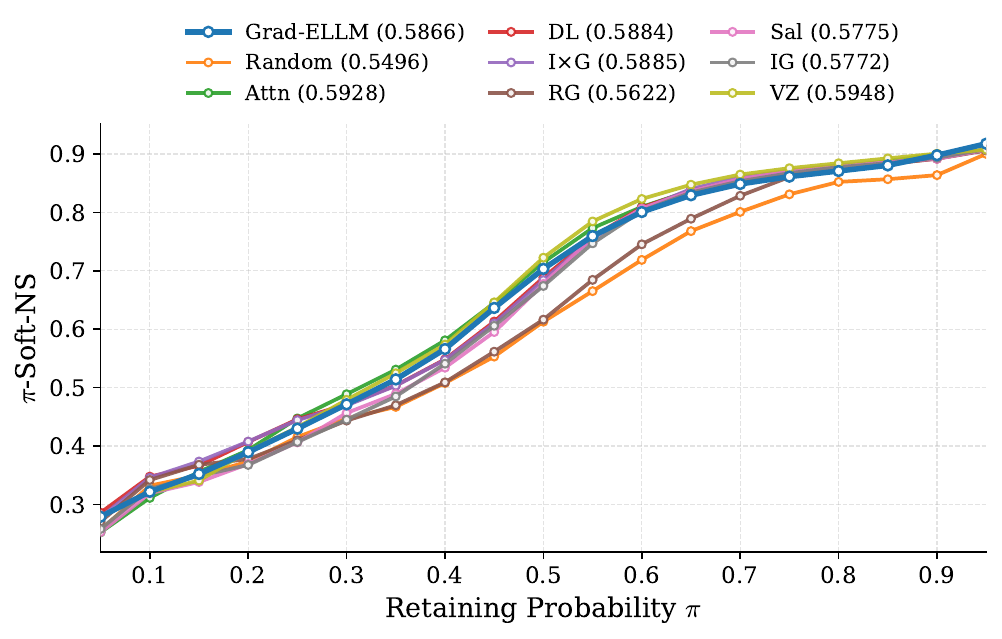}\\[-0.8em]
    {\small (a) $\pi$-Soft-NS vs. Retaining Probability}
  \end{minipage}
  \hfill
  \begin{minipage}[t]{0.485\textwidth}
    \centering
    \includegraphics[width=\linewidth]{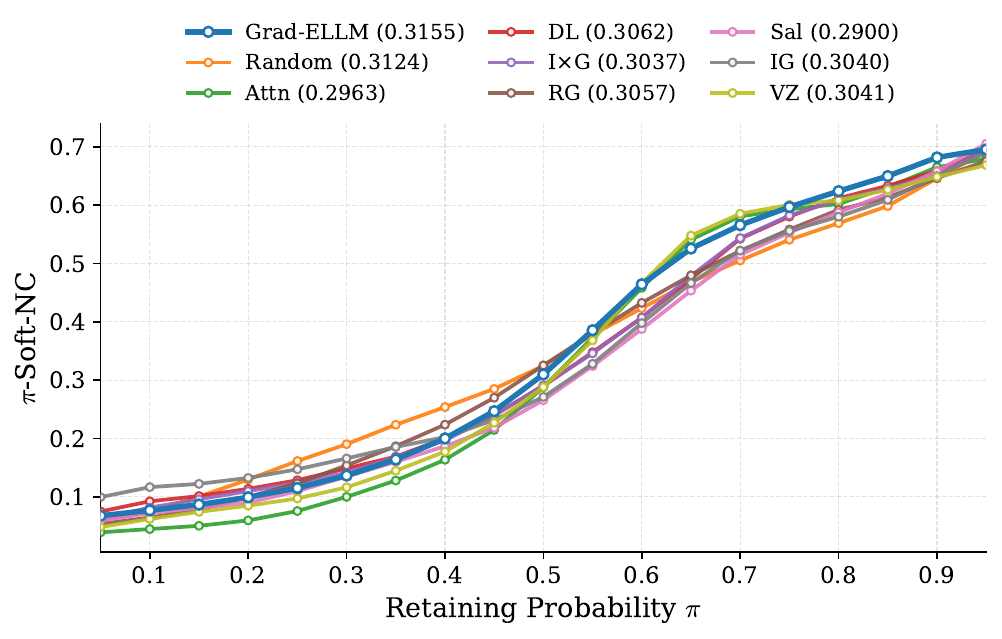}\\[-0.8em]
    {\small (b) $\pi$-Soft-NC vs. Retaining Probability}
  \end{minipage}

  \vspace{-0.6em}
  \caption{Proposed $\pi$-Soft-NC/NS vs. Retaining Probability on SST2 for different XAI methods with Llama.}
  \label{soft-ns/nc_fig}
  \vspace{-1.0em}
\end{figure*}

\paragraph{Diagnosis of original Soft-NC/NS.}
Before comparing methods under the proposed metrics, we first examine the original Soft-NC/NS of \citet{zhao2024reagent}. As shown in Tab.~\ref{llama verify}, raw Soft-NC/NS compares methods under substantially different expected numbers of retained tokens. For example, Random and Grad-ELLM retain many more tokens on average ($\mathbb{E}[R]=80$ and $62$) than sparse gradient-based methods such as DeepLIFT, IG, and Saliency ($\mathbb{E}[R]\approx18$--$21$). In other words, such evaluation is like erroneously comparing results with retaining probability $\pi=0.1$ to those with $\pi=0.4$ in Fig.~\ref{soft-ns/nc_fig}.
Therefore, higher Soft-NC/NS scores reflect more words involved in the perturbation rather than more faithful attribution, confirming the need to control the retaining probability.

\begin{table}[htbp]
  \centering
  \tiny
  \setlength{\tabcolsep}{3.5pt}
  \begin{tabular}{ccccccccc}
    \hline
    \ Metric&Attn&DL&I$\times$G&IG&Rnd&Sal&VZ&Ours \\
    \hline
    Soft-NS$\uparrow$& 0.052 & 0.087 & 0.086 & 0.089 & 0.132 & 0.087 & 0.032 & 0.138 \\
    Soft-NC$\uparrow$& 0.004 & 0.006 & 0.006 & 0.005 & 0.005 & 0.006 & 0.004 & 0.013 \\
    \hline
     $\mathbb{E}[R]$& 17 & 20 & 19 & 18 & 80 & 21 & 8 & 62 \\
    \hline
  \end{tabular}
  \caption{\label{llama verify}
    Faithfulness evaluation of \emph{original} Soft-NC/NS on IMDb with Llama. No transformations are applied to the attribution scores. $\mathbb{E}[R]$ denotes the expected number of retained words. The abbreviations are the same as in Table \ref{tab:soft-faithfulness-sidebyside}.
  }
\end{table}

\paragraph{Evaluation under $\pi$-Soft-NC/NS.}
Tab.~\ref{tab:soft-faithfulness-sidebyside} reports the AUC of $\pi$-Soft-NC/NS curves. 
Grad-ELLM achieves the best average $\pi$-Soft-NC on both Llama and Mistral, indicating strong comprehensiveness-oriented faithfulness. On $\pi$-Soft-NS, no clear method dominates, with only 
Input$\times$Gradient in the top-3 for both models. 
Fig.~\ref{soft-ns/nc_fig} shows an example set of curves on one dataset. For $\pi$-Soft-NS, most methods see advantage over Random in the range $\pi\in[0.5,0.65]$. For $\pi$-Soft-NC, some methods see advantage over Random after when $\pi \geq 0.6$ .

\paragraph{Insights}
Our results in Tab.~\ref{tab:soft-faithfulness-sidebyside} reveal a sufficiency-comprehensiveness asymmetry: 
Grad-ELLM is strongest on $\pi$-Soft-NC but not always on $\pi$-Soft-NS, indicating stronger broad evidence coverage than precise orderings of the top attributions.  Meanwhile, sparse attribution methods, like Input$\times$Gradient perform well on $\pi$-Soft-NS, indicating that they more precisely identify the most important words. However, it should be noted that the attribution methods only show advantage over Random attributions for larger retaining probability $\pi>0.5$. 
It is unclear whether this is due to the need of contextual information to perform the task (i.e., more words need to be retained), or that the top-ranked attributions are misordered.

Finally, we further examine whether the proposed $\pi$-Soft metrics merely favor dense attribution maps in Appendix~\ref{app:attribution-distribution}. The attribution-value distributions show that density alone is insufficient: Random produces the most uniform scores but performs poorly under $\pi$-Soft metrics. This suggests that $\pi$-Soft-NC/NS reward calibrated selectivity rather than density alone.

\begin{figure}[htb]
  \centering

  \begin{minipage}[t]{0.96\linewidth}
    \centering
    \includegraphics[width=\linewidth]{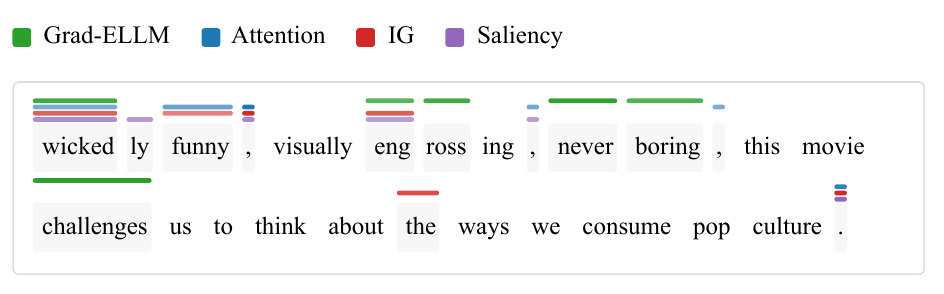}\\[-0.3em]
    \small (a) Positive SST2 example
  \end{minipage}

  \vspace{0.8em}

  \begin{minipage}[t]{0.96\linewidth}
    \centering
    \includegraphics[width=\linewidth]{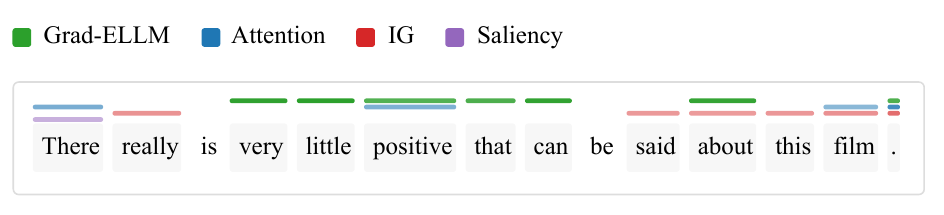}\\[-0.3em]
    \small (b) Negative IMDb example
  \end{minipage}

  \caption{
Qualitative comparison of token-level attributions on sentiment classification examples.
Colored bars above tokens indicate the top 10\% attribution scores by each method.
For the long IMDb review, we show the key sentence containing the decisive sentiment phrase.}
\label{fig:qualitative_sentiment_examples}
\end{figure}

\subsubsection{Qualitative Results}
Fig.~\ref{fig:qualitative_sentiment_examples} provides qualitative examples of token-level attributions on sentiment classification tasks. 
For readability, we visualize whether each token is selected among the top 10\% attribution scores by each method, rather than showing continuous attribution values. 
The selected methods are compared on the same input, and colored bars above each token indicate the tokens selected by different attribution methods.

In the positive example, Grad-ELLM selects sentiment-bearing tokens and phrases such as ``engrossing'' and ``never boring'', which jointly support the positive prediction. 
Attention and gradient-based baselines also select some relevant tokens, but their selected tokens are less consistently aligned with the main positive sentiment cues.

In the negative example, the review is much longer, so we show the key sentence containing the decisive sentiment phrase. 
Grad-ELLM highlights the phrase ``very little positive'', which is crucial because the word ``positive'' alone is misleading without the preceding negation-like modifier. 
In contrast, other methods tend to focus on isolated tokens such as ``positive'' or nearby tokens, and may miss the compositional phrase that explains the negative prediction. 
The full example and additional examples are provided in Appendix~\ref{app:qualitative_full_examples}.

\section{Conclusion}

In this work, we studied faithfulness evaluation for decoder-only LLM attributions. We first identified a 
bias in the existing Soft-NC/NS metrics \citep{zhao2024reagent}, where different attribution score distributions can lead to different expected amounts of retained input information. To address this issue, we proposed $\pi$-Soft-NC and $\pi$-Soft-NS, which compare attribution methods under a controlled target retaining probability. We further introduced Grad-ELLM, a gradient-based attribution method that combines gradient-derived channel importance with attention-derived token importance for autoregressive next-token prediction. Experiments across classification and generation tasks show that Grad-ELLM achieves strong comprehensiveness-oriented faithfulness under $\pi$-Soft-NC. Meanwhile, there is no clearly dominant method when evaluating with the $\pi$-Soft-NS.
Compared to Random attributions, most methods only see advantage for larger retaining probability ($\pi>0.5$), which indicates gaps in both the attribution methods themselves and our understanding of the role of context in the task.
We hope that our proposed faithfulness metric and protocol provide a rigorous framework for evaluating XAI attribution methods for LLMs, leading to future progress in the field.

\clearpage

\section{Limitations}
Grad-ELLM is designed to produce calibrated attribution distributions rather than extremely sparse token rankings. This design is well aligned with soft distributional metrics such as $\pi$-Soft-NC/NS, but it may be less aligned with ranking-based hard perturbation metrics such as insertion/deletion \citep{DBLP:conf/bmvc/PetsiukDS18}, where the ordering of a few top-ranked tokens dominates the final AUC. Therefore, Grad-ELLM should not be interpreted as universally optimal under all faithfulness metrics; instead, its strength lies in providing a more complete and calibrated attribution distribution.


Furthermore, our study focuses exclusively on faithfulness and does not include evaluations of plausibility, which assesses how intuitive and human-understandable the explanations are. Plausibility is inherently subjective, and quantifying it poses significant challenges—particularly when comparing dense heatmaps (which provide a holistic view of token contributions) to sparse ones (which emphasize focal points). Without standardized, objective metrics or large-scale user studies, it is difficult to determine which type offers better interpretability in practice.

Grad-ELLM also relies on access to internal model states, such as gradients and attention layers, restricting its use to open-source LLMs like Llama and Mistral. Extending it to proprietary models or integrating it with global interpretability techniques in future research could address these gaps.

Another limitation is that $\pi$-Soft-NC/NS still rely on embedding-level perturbations. Although controlling the retaining probability improves fairness across attribution methods, the perturbed representations may still differ from naturally occurring text inputs, which is a general concern in perturbation-based interpretability \citep{ancona2018towards,bastings-filippova-2020-elephant,yin2022sensitivity}. Future work could combine retained-information control with more semantically constrained perturbations.

\bibliography{main}

@article{esteva2019guide,
  title={A guide to deep learning in healthcare},
  author={Esteva, Andre and Robicquet, Alexandre and Ramsundar, Bharath and Kuleshov, Volodymyr and DePristo, Mark and Chou, Katherine and Cui, Claire and Corrado, Greg and Thrun, Sebastian and Dean, Jeff},
  journal={Nature medicine},
  volume={25},
  number={1},
  pages={24--29},
  year={2019},
  publisher={Nature Publishing Group US New York}
}

@article{veale2017fairer,
  title={Fairer machine learning in the real world: Mitigating discrimination without collecting sensitive data},
  author={Veale, Michael and Binns, Reuben},
  journal={Big Data \& Society},
  volume={4},
  number={2},
  pages={2053951717743530},
  year={2017},
  publisher={SAGE Publications Sage UK: London, England}
}

@misc{grattafiori2024llama3herdmodels,
      title={The Llama 3 Herd of Models}, 
      author={Aaron Grattafiori and Abhimanyu Dubey and Abhinav Jauhri and Abhinav Pandey and Abhishek Kadian and Ahmad Al-Dahle and Aiesha Letman and Akhil Mathur and Alan Schelten and Alex Vaughan and Amy Yang and Angela Fan and Anirudh Goyal and Anthony Hartshorn and Aobo Yang and Archi Mitra and Archie Sravankumar and Artem Korenev and Arthur Hinsvark and Arun Rao and Aston Zhang and Aurelien Rodriguez and Austen Gregerson and Ava Spataru and Baptiste Roziere and Bethany Biron and Binh Tang and Bobbie Chern and Charlotte Caucheteux and Chaya Nayak and Chloe Bi and Chris Marra and Chris McConnell and Christian Keller and Christophe Touret and Chunyang Wu and Corinne Wong and Cristian Canton Ferrer and Cyrus Nikolaidis and Damien Allonsius and Daniel Song and Danielle Pintz and Danny Livshits and Danny Wyatt and David Esiobu and Dhruv Choudhary and Dhruv Mahajan and Diego Garcia-Olano and Diego Perino and Dieuwke Hupkes and Egor Lakomkin and Ehab AlBadawy and Elina Lobanova and Emily Dinan and Eric Michael Smith and Filip Radenovic and Francisco Guzmán and Frank Zhang and Gabriel Synnaeve and Gabrielle Lee and Georgia Lewis Anderson and Govind Thattai and Graeme Nail and Gregoire Mialon and Guan Pang and Guillem Cucurell and Hailey Nguyen and Hannah Korevaar and Hu Xu and Hugo Touvron and Iliyan Zarov and Imanol Arrieta Ibarra and Isabel Kloumann and Ishan Misra and Ivan Evtimov and Jack Zhang and Jade Copet and Jaewon Lee and Jan Geffert and Jana Vranes and Jason Park and Jay Mahadeokar and Jeet Shah and Jelmer van der Linde and Jennifer Billock and Jenny Hong and Jenya Lee and Jeremy Fu and Jianfeng Chi and Jianyu Huang and Jiawen Liu and Jie Wang and Jiecao Yu and Joanna Bitton and Joe Spisak and Jongsoo Park and Joseph Rocca and Joshua Johnstun and Joshua Saxe and Junteng Jia and Kalyan Vasuden Alwala and Karthik Prasad and Kartikeya Upasani and Kate Plawiak and Ke Li and Kenneth Heafield and Kevin Stone and Khalid El-Arini and Krithika Iyer and Kshitiz Malik and Kuenley Chiu and Kunal Bhalla and Kushal Lakhotia and Lauren Rantala-Yeary and Laurens van der Maaten and Lawrence Chen and Liang Tan and Liz Jenkins and Louis Martin and Lovish Madaan and Lubo Malo and Lukas Blecher and Lukas Landzaat and Luke de Oliveira and Madeline Muzzi and Mahesh Pasupuleti and Mannat Singh and Manohar Paluri and Marcin Kardas and Maria Tsimpoukelli and Mathew Oldham and Mathieu Rita and Maya Pavlova and Melanie Kambadur and Mike Lewis and Min Si and Mitesh Kumar Singh and Mona Hassan and Naman Goyal and Narjes Torabi and Nikolay Bashlykov and Nikolay Bogoychev and Niladri Chatterji and Ning Zhang and Olivier Duchenne and Onur Çelebi and Patrick Alrassy and Pengchuan Zhang and Pengwei Li and Petar Vasic and Peter Weng and Prajjwal Bhargava and Pratik Dubal and Praveen Krishnan and Punit Singh Koura and Puxin Xu and Qing He and Qingxiao Dong and Ragavan Srinivasan and Raj Ganapathy and Ramon Calderer and Ricardo Silveira Cabral and Robert Stojnic and Roberta Raileanu and Rohan Maheswari and Rohit Girdhar and Rohit Patel and Romain Sauvestre and Ronnie Polidoro and Roshan Sumbaly and Ross Taylor and Ruan Silva and Rui Hou and Rui Wang and Saghar Hosseini and Sahana Chennabasappa and Sanjay Singh and Sean Bell and Seohyun Sonia Kim and Sergey Edunov and Shaoliang Nie and Sharan Narang and Sharath Raparthy and Sheng Shen and Shengye Wan and Shruti Bhosale and Shun Zhang and Simon Vandenhende and Soumya Batra and Spencer Whitman and Sten Sootla and Stephane Collot and Suchin Gururangan and Sydney Borodinsky and Tamar Herman and Tara Fowler and Tarek Sheasha and Thomas Georgiou and Thomas Scialom and Tobias Speckbacher and Todor Mihaylov and Tong Xiao and Ujjwal Karn and Vedanuj Goswami and Vibhor Gupta and Vignesh Ramanathan and Viktor Kerkez and Vincent Gonguet and Virginie Do and Vish Vogeti and Vítor Albiero and Vladan Petrovic and Weiwei Chu and Wenhan Xiong and Wenyin Fu and Whitney Meers and Xavier Martinet and Xiaodong Wang and Xiaofang Wang and Xiaoqing Ellen Tan and Xide Xia and Xinfeng Xie and Xuchao Jia and Xuewei Wang and Yaelle Goldschlag and Yashesh Gaur and Yasmine Babaei and Yi Wen and Yiwen Song and Yuchen Zhang and Yue Li and Yuning Mao and Zacharie Delpierre Coudert and Zheng Yan and Zhengxing Chen and Zoe Papakipos and Aaditya Singh and Aayushi Srivastava and Abha Jain and Adam Kelsey and Adam Shajnfeld and Adithya Gangidi and Adolfo Victoria and Ahuva Goldstand and Ajay Menon and Ajay Sharma and Alex Boesenberg and Alexei Baevski and Allie Feinstein and Amanda Kallet and Amit Sangani and Amos Teo and Anam Yunus and Andrei Lupu and Andres Alvarado and Andrew Caples and Andrew Gu and Andrew Ho and Andrew Poulton and Andrew Ryan and Ankit Ramchandani and Annie Dong and Annie Franco and Anuj Goyal and Aparajita Saraf and Arkabandhu Chowdhury and Ashley Gabriel and Ashwin Bharambe and Assaf Eisenman and Azadeh Yazdan and Beau James and Ben Maurer and Benjamin Leonhardi and Bernie Huang and Beth Loyd and Beto De Paola and Bhargavi Paranjape and Bing Liu and Bo Wu and Boyu Ni and Braden Hancock and Bram Wasti and Brandon Spence and Brani Stojkovic and Brian Gamido and Britt Montalvo and Carl Parker and Carly Burton and Catalina Mejia and Ce Liu and Changhan Wang and Changkyu Kim and Chao Zhou and Chester Hu and Ching-Hsiang Chu and Chris Cai and Chris Tindal and Christoph Feichtenhofer and Cynthia Gao and Damon Civin and Dana Beaty and Daniel Kreymer and Daniel Li and David Adkins and David Xu and Davide Testuggine and Delia David and Devi Parikh and Diana Liskovich and Didem Foss and Dingkang Wang and Duc Le and Dustin Holland and Edward Dowling and Eissa Jamil and Elaine Montgomery and Eleonora Presani and Emily Hahn and Emily Wood and Eric-Tuan Le and Erik Brinkman and Esteban Arcaute and Evan Dunbar and Evan Smothers and Fei Sun and Felix Kreuk and Feng Tian and Filippos Kokkinos and Firat Ozgenel and Francesco Caggioni and Frank Kanayet and Frank Seide and Gabriela Medina Florez and Gabriella Schwarz and Gada Badeer and Georgia Swee and Gil Halpern and Grant Herman and Grigory Sizov and Guangyi and Zhang and Guna Lakshminarayanan and Hakan Inan and Hamid Shojanazeri and Han Zou and Hannah Wang and Hanwen Zha and Haroun Habeeb and Harrison Rudolph and Helen Suk and Henry Aspegren and Hunter Goldman and Hongyuan Zhan and Ibrahim Damlaj and Igor Molybog and Igor Tufanov and Ilias Leontiadis and Irina-Elena Veliche and Itai Gat and Jake Weissman and James Geboski and James Kohli and Janice Lam and Japhet Asher and Jean-Baptiste Gaya and Jeff Marcus and Jeff Tang and Jennifer Chan and Jenny Zhen and Jeremy Reizenstein and Jeremy Teboul and Jessica Zhong and Jian Jin and Jingyi Yang and Joe Cummings and Jon Carvill and Jon Shepard and Jonathan McPhie and Jonathan Torres and Josh Ginsburg and Junjie Wang and Kai Wu and Kam Hou U and Karan Saxena and Kartikay Khandelwal and Katayoun Zand and Kathy Matosich and Kaushik Veeraraghavan and Kelly Michelena and Keqian Li and Kiran Jagadeesh and Kun Huang and Kunal Chawla and Kyle Huang and Lailin Chen and Lakshya Garg and Lavender A and Leandro Silva and Lee Bell and Lei Zhang and Liangpeng Guo and Licheng Yu and Liron Moshkovich and Luca Wehrstedt and Madian Khabsa and Manav Avalani and Manish Bhatt and Martynas Mankus and Matan Hasson and Matthew Lennie and Matthias Reso and Maxim Groshev and Maxim Naumov and Maya Lathi and Meghan Keneally and Miao Liu and Michael L. Seltzer and Michal Valko and Michelle Restrepo and Mihir Patel and Mik Vyatskov and Mikayel Samvelyan and Mike Clark and Mike Macey and Mike Wang and Miquel Jubert Hermoso and Mo Metanat and Mohammad Rastegari and Munish Bansal and Nandhini Santhanam and Natascha Parks and Natasha White and Navyata Bawa and Nayan Singhal and Nick Egebo and Nicolas Usunier and Nikhil Mehta and Nikolay Pavlovich Laptev and Ning Dong and Norman Cheng and Oleg Chernoguz and Olivia Hart and Omkar Salpekar and Ozlem Kalinli and Parkin Kent and Parth Parekh and Paul Saab and Pavan Balaji and Pedro Rittner and Philip Bontrager and Pierre Roux and Piotr Dollar and Polina Zvyagina and Prashant Ratanchandani and Pritish Yuvraj and Qian Liang and Rachad Alao and Rachel Rodriguez and Rafi Ayub and Raghotham Murthy and Raghu Nayani and Rahul Mitra and Rangaprabhu Parthasarathy and Raymond Li and Rebekkah Hogan and Robin Battey and Rocky Wang and Russ Howes and Ruty Rinott and Sachin Mehta and Sachin Siby and Sai Jayesh Bondu and Samyak Datta and Sara Chugh and Sara Hunt and Sargun Dhillon and Sasha Sidorov and Satadru Pan and Saurabh Mahajan and Saurabh Verma and Seiji Yamamoto and Sharadh Ramaswamy and Shaun Lindsay and Shaun Lindsay and Sheng Feng and Shenghao Lin and Shengxin Cindy Zha and Shishir Patil and Shiva Shankar and Shuqiang Zhang and Shuqiang Zhang and Sinong Wang and Sneha Agarwal and Soji Sajuyigbe and Soumith Chintala and Stephanie Max and Stephen Chen and Steve Kehoe and Steve Satterfield and Sudarshan Govindaprasad and Sumit Gupta and Summer Deng and Sungmin Cho and Sunny Virk and Suraj Subramanian and Sy Choudhury and Sydney Goldman and Tal Remez and Tamar Glaser and Tamara Best and Thilo Koehler and Thomas Robinson and Tianhe Li and Tianjun Zhang and Tim Matthews and Timothy Chou and Tzook Shaked and Varun Vontimitta and Victoria Ajayi and Victoria Montanez and Vijai Mohan and Vinay Satish Kumar and Vishal Mangla and Vlad Ionescu and Vlad Poenaru and Vlad Tiberiu Mihailescu and Vladimir Ivanov and Wei Li and Wenchen Wang and Wenwen Jiang and Wes Bouaziz and Will Constable and Xiaocheng Tang and Xiaojian Wu and Xiaolan Wang and Xilun Wu and Xinbo Gao and Yaniv Kleinman and Yanjun Chen and Ye Hu and Ye Jia and Ye Qi and Yenda Li and Yilin Zhang and Ying Zhang and Yossi Adi and Youngjin Nam and Yu and Wang and Yu Zhao and Yuchen Hao and Yundi Qian and Yunlu Li and Yuzi He and Zach Rait and Zachary DeVito and Zef Rosnbrick and Zhaoduo Wen and Zhenyu Yang and Zhiwei Zhao and Zhiyu Ma},
      year={2024},
      eprint={2407.21783},
      archivePrefix={arXiv},
      primaryClass={cs.AI},
      url={https://arxiv.org/abs/2407.21783}, 
}

@inproceedings{deyoung-etal-2020-eraser,
    title = "{ERASER}: {A} Benchmark to Evaluate Rationalized {NLP} Models",
    author = "DeYoung, Jay  and
      Jain, Sarthak  and
      Rajani, Nazneen Fatema  and
      Lehman, Eric  and
      Xiong, Caiming  and
      Socher, Richard  and
      Wallace, Byron C.",
    editor = "Jurafsky, Dan  and
      Chai, Joyce  and
      Schluter, Natalie  and
      Tetreault, Joel",
    booktitle = "Proceedings of the 58th Annual Meeting of the Association for Computational Linguistics",
    month = jul,
    year = "2020",
    address = "Online",
    publisher = "Association for Computational Linguistics",
    url = "https://aclanthology.org/2020.acl-main.408/",
    doi = "10.18653/v1/2020.acl-main.408",
    pages = "4443--4458"
}

@inproceedings{sundararajan2017axiomatic,
  title={Axiomatic attribution for deep networks},
  author={Sundararajan, Mukund and Taly, Ankur and Yan, Qiqi},
  booktitle={International conference on machine learning},
  pages={3319--3328},
  year={2017},
  organization={PMLR}
}

@inproceedings{lei-etal-2016-rationalizing,
    title = "Rationalizing Neural Predictions",
    author = "Lei, Tao  and
      Barzilay, Regina  and
      Jaakkola, Tommi",
    editor = "Su, Jian  and
      Duh, Kevin  and
      Carreras, Xavier",
    booktitle = "Proceedings of the 2016 Conference on Empirical Methods in Natural Language Processing",
    month = nov,
    year = "2016",
    address = "Austin, Texas",
    publisher = "Association for Computational Linguistics",
    url = "https://aclanthology.org/D16-1011/",
    doi = "10.18653/v1/D16-1011",
    pages = "107--117"
}

@inproceedings{ribeiro2016should,
  title={" Why should i trust you?" Explaining the predictions of any classifier},
  author={Ribeiro, Marco Tulio and Singh, Sameer and Guestrin, Carlos},
  booktitle={Proceedings of the 22nd ACM SIGKDD international conference on knowledge discovery and data mining},
  pages={1135--1144},
  year={2016}
}

@inproceedings{DBLP:journals/corr/SimonyanVZ13,
  author       = {Karen Simonyan and
                  Andrea Vedaldi and
                  Andrew Zisserman},
  editor       = {Yoshua Bengio and
                  Yann LeCun},
  title        = {Deep Inside Convolutional Networks: Visualising Image Classification
                  Models and Saliency Maps},
  booktitle    = {2nd International Conference on Learning Representations, {ICLR} 2014,
                  Banff, AB, Canada, April 14-16, 2014, Workshop Track Proceedings},
  year         = {2014},
  url          = {http://arxiv.org/abs/1312.6034},
  bibsource    = {dblp computer science bibliography, https://dblp.org}
}

@inproceedings{shrikumar2017learning,
  title={Learning important features through propagating activation differences},
  author={Shrikumar, Avanti and Greenside, Peyton and Kundaje, Anshul},
  booktitle={International conference on machine learning},
  pages={3145--3153},
  year={2017},
  organization={PMlR}
}

@inproceedings{mohebbi-etal-2023-quantifying,
    title = "Quantifying Context Mixing in Transformers",
    author = "Mohebbi, Hosein  and
      Zuidema, Willem  and
      Chrupa{\l}a, Grzegorz  and
      Alishahi, Afra",
    editor = "Vlachos, Andreas  and
      Augenstein, Isabelle",
    booktitle = "Proceedings of the 17th Conference of the European Chapter of the Association for Computational Linguistics",
    month = may,
    year = "2023",
    address = "Dubrovnik, Croatia",
    publisher = "Association for Computational Linguistics",
    url = "https://aclanthology.org/2023.eacl-main.245/",
    doi = "10.18653/v1/2023.eacl-main.245",
    pages = "3378--3400"
}

@misc{fayyaz2024evaluatinghumanalignmentmodel,
      title={Evaluating Human Alignment and Model Faithfulness of LLM Rationale}, 
      author={Mohsen Fayyaz and Fan Yin and Jiao Sun and Nanyun Peng},
      year={2024},
      eprint={2407.00219},
      archivePrefix={arXiv},
      primaryClass={cs.CL},
      url={https://arxiv.org/abs/2407.00219}, 
}

@inproceedings{sarti-etal-2023-inseq,
    title = "Inseq: An Interpretability Toolkit for Sequence Generation Models",
    author = "Sarti, Gabriele  and
      Feldhus, Nils  and
      Sickert, Ludwig  and
      van der Wal, Oskar and
      Nissim, Malvina and
      Bisazza, Arianna",
    booktitle = "Proceedings of the 61st Annual Meeting of the Association for Computational Linguistics (Volume 3: System Demonstrations)",
    month = jul,
    year = "2023",
    address = "Toronto, Canada",
    publisher = "Association for Computational Linguistics",
    url = "https://aclanthology.org/2023.acl-demo.40",
    doi = "10.18653/v1/2023.acl-demo.40",
    pages = "421--435",
}

@inproceedings{DBLP:journals/corr/BahdanauCB14,
  author       = {Dzmitry Bahdanau and
                  Kyunghyun Cho and
                  Yoshua Bengio},
  editor       = {Yoshua Bengio and
                  Yann LeCun},
  title        = {Neural Machine Translation by Jointly Learning to Align and Translate},
  booktitle    = {3rd International Conference on Learning Representations, {ICLR} 2015,
                  San Diego, CA, USA, May 7-9, 2015, Conference Track Proceedings},
  year         = {2015},
  url          = {http://arxiv.org/abs/1409.0473},
  bibsource    = {dblp computer science bibliography, https://dblp.org}
}

@article{zhao2024reagent,
  title={Reagent: A model-agnostic feature attribution method for generative language models},
  author={Zhao, Zhixue and Shan, Boxuan},
  journal={arXiv preprint arXiv:2402.00794},
  year={2024}
}

@inproceedings{zhao2024gradient,
  title={Gradient-based visual explanation for transformer-based clip},
  author={Zhao, Chenyang and Wang, Kun and Zeng, Xingyu and Zhao, Rui and Chan, Antoni B},
  booktitle={International Conference on Machine Learning},
  pages={61072--61091},
  year={2024},
  organization={PMLR}
}

@inproceedings{radford2021learning,
  title={Learning transferable visual models from natural language supervision},
  author={Radford, Alec and Kim, Jong Wook and Hallacy, Chris and Ramesh, Aditya and Goh, Gabriel and Agarwal, Sandhini and Sastry, Girish and Askell, Amanda and Mishkin, Pamela and Clark, Jack and others},
  booktitle={International conference on machine learning},
  pages={8748--8763},
  year={2021},
  organization={PmLR}
}

@article{zhao2024explainability,
  title={Explainability for large language models: A survey},
  author={Zhao, Haiyan and Chen, Hanjie and Yang, Fan and Liu, Ninghao and Deng, Huiqi and Cai, Hengyi and Wang, Shuaiqiang and Yin, Dawei and Du, Mengnan},
  journal={ACM Transactions on Intelligent Systems and Technology},
  volume={15},
  number={2},
  pages={1--38},
  year={2024},
  publisher={ACM New York, NY}
}

@article{luo2024understanding,
  title={From understanding to utilization: A survey on explainability for large language models},
  author={Luo, Haoyan and Specia, Lucia},
  journal={arXiv preprint arXiv:2401.12874},
  year={2024}
}

@inproceedings{maas2011learning,
  title={Learning word vectors for sentiment analysis},
  author={Maas, Andrew and Daly, Raymond E and Pham, Peter T and Huang, Dan and Ng, Andrew Y and Potts, Christopher},
  booktitle={Proceedings of the 49th annual meeting of the association for computational linguistics: Human language technologies},
  pages={142--150},
  year={2011}
}

@inproceedings{socher2013recursive,
  title={Recursive deep models for semantic compositionality over a sentiment treebank},
  author={Socher, Richard and Perelygin, Alex and Wu, Jean and Chuang, Jason and Manning, Christopher D and Ng, Andrew Y and Potts, Christopher},
  booktitle={Proceedings of the 2013 conference on empirical methods in natural language processing},
  pages={1631--1642},
  year={2013}
}

@inproceedings{lal-etal-2021-tellmewhy,
    title = "{T}ell{M}e{W}hy: A Dataset for Answering Why-Questions in Narratives",
    author = "Lal, Yash Kumar  and
      Chambers, Nathanael  and
      Mooney, Raymond  and
      Balasubramanian, Niranjan",
    editor = "Zong, Chengqing  and
      Xia, Fei  and
      Li, Wenjie  and
      Navigli, Roberto",
    booktitle = "Findings of the Association for Computational Linguistics: ACL-IJCNLP 2021",
    month = aug,
    year = "2021",
    address = "Online",
    publisher = "Association for Computational Linguistics",
    url = "https://aclanthology.org/2021.findings-acl.53/",
    doi = "10.18653/v1/2021.findings-acl.53",
    pages = "596--610"
}

@misc{liu2019robertarobustlyoptimizedbert,
      title={RoBERTa: A Robustly Optimized BERT Pretraining Approach}, 
      author={Yinhan Liu and Myle Ott and Naman Goyal and Jingfei Du and Mandar Joshi and Danqi Chen and Omer Levy and Mike Lewis and Luke Zettlemoyer and Veselin Stoyanov},
      year={2019},
      eprint={1907.11692},
      archivePrefix={arXiv},
      primaryClass={cs.CL},
      url={https://arxiv.org/abs/1907.11692}, 
}

@article{vaswani2017attention,
  title={Attention is all you need},
  author={Vaswani, Ashish and Shazeer, Noam and Parmar, Niki and Uszkoreit, Jakob and Jones, Llion and Gomez, Aidan N and Kaiser, {\L}ukasz and Polosukhin, Illia},
  journal={Advances in neural information processing systems},
  volume={30},
  year={2017}
}

@inproceedings{barkan2021grad,
  title={Grad-sam: Explaining transformers via gradient self-attention maps},
  author={Barkan, Oren and Hauon, Edan and Caciularu, Avi and Katz, Ori and Malkiel, Itzik and Armstrong, Omri and Koenigstein, Noam},
  booktitle={Proceedings of the 30th ACM International Conference on Information \& Knowledge Management},
  pages={2882--2887},
  year={2021}
}

@inproceedings{DBLP:conf/bmvc/PetsiukDS18,
  author       = {Vitali Petsiuk and
                  Abir Das and
                  Kate Saenko},
  title        = {{RISE:} Randomized Input Sampling for Explanation of Black-box Models},
  booktitle    = {British Machine Vision Conference 2018, {BMVC} 2018, Newcastle, UK,
                  September 3-6, 2018},
  pages        = {151},
  publisher    = {{BMVA} Press},
  year         = {2018},
  url          = {http://bmvc2018.org/contents/papers/1064.pdf},
  bibsource    = {dblp computer science bibliography, https://dblp.org}
}

@inproceedings{manakul2023selfcheckgpt,
  title={Selfcheckgpt: Zero-resource black-box hallucination detection for generative large language models},
  author={Manakul, Potsawee and Liusie, Adian and Gales, Mark},
  booktitle={Proceedings of the 2023 conference on empirical methods in natural language processing},
  pages={9004--9017},
  year={2023}
}

@misc{jiang2023mistral7b,
      title={Mistral 7B}, 
      author={Albert Q. Jiang and Alexandre Sablayrolles and Arthur Mensch and Chris Bamford and Devendra Singh Chaplot and Diego de las Casas and Florian Bressand and Gianna Lengyel and Guillaume Lample and Lucile Saulnier and Lélio Renard Lavaud and Marie-Anne Lachaux and Pierre Stock and Teven Le Scao and Thibaut Lavril and Thomas Wang and Timothée Lacroix and William El Sayed},
      year={2023},
      eprint={2310.06825},
      archivePrefix={arXiv},
      primaryClass={cs.CL},
      url={https://arxiv.org/abs/2310.06825}, 
}

@inproceedings{wolf2020transformers,
  title={Transformers: State-of-the-art natural language processing},
  author={Wolf, Thomas and Debut, Lysandre and Sanh, Victor and Chaumond, Julien and Delangue, Clement and Moi, Anthony and Cistac, Pierric and Rault, Tim and Louf, Remi and Funtowicz, Morgan and others},
  booktitle={Proceedings of the 2020 conference on empirical methods in natural language processing: system demonstrations},
  pages={38--45},
  year={2020}
}

@inproceedings{Jain2019AttentionIN,
  title={Attention is not Explanation},
  author={Sarthak Jain and Byron C. Wallace},
  booktitle={North American Chapter of the Association for Computational Linguistics},
  year={2019},
  url={https://api.semanticscholar.org/CorpusID:67855860}
}

@inproceedings{serrano-smith-2019-attention,
    title = "Is Attention Interpretable?",
    author = "Serrano, Sofia  and
      Smith, Noah A.",
    editor = "Korhonen, Anna  and
      Traum, David  and
      M{\`a}rquez, Llu{\'i}s",
    booktitle = "Proceedings of the 57th Annual Meeting of the Association for Computational Linguistics",
    month = jul,
    year = "2019",
    address = "Florence, Italy",
    publisher = "Association for Computational Linguistics",
    url = "https://aclanthology.org/P19-1282/",
    doi = "10.18653/v1/P19-1282",
    pages = "2931--2951"
}

@inproceedings{ancona2018towards,
  title={Towards better understanding of gradient-based attribution methods for Deep Neural Networks},
  author={Ancona, Marco and Ceolini, Enea and {\"O}ztireli, Cengiz and Gross, Markus},
  booktitle={International Conference on Learning Representations},
  year={2018}
}

@inproceedings{bastings-filippova-2020-elephant,
    title = "The elephant in the interpretability room: Why use attention as explanation when we have saliency methods?",
    author = "Bastings, Jasmijn  and
      Filippova, Katja",
    editor = "Alishahi, Afra  and
      Belinkov, Yonatan  and
      Chrupa{\l}a, Grzegorz  and
      Hupkes, Dieuwke  and
      Pinter, Yuval  and
      Sajjad, Hassan",
    booktitle = "Proceedings of the Third BlackboxNLP Workshop on Analyzing and Interpreting Neural Networks for NLP",
    month = nov,
    year = "2020",
    address = "Online",
    publisher = "Association for Computational Linguistics",
    url = "https://aclanthology.org/2020.blackboxnlp-1.14/",
    doi = "10.18653/v1/2020.blackboxnlp-1.14",
    pages = "149--155"
}

@inproceedings{yin2022sensitivity,
  title={On the sensitivity and stability of model interpretations in NLP},
  author={Yin, Fan and Shi, Zhouxing and Hsieh, Cho-Jui and Chang, Kai-Wei},
  booktitle={Proceedings of the 60th Annual Meeting of the Association for Computational Linguistics (Volume 1: Long Papers)},
  pages={2631--2647},
  year={2022}
}

@article{samek2016evaluating,
  title={Evaluating the visualization of what a deep neural network has learned},
  author={Samek, Wojciech and Binder, Alexander and Montavon, Gr{\'e}goire and Lapuschkin, Sebastian and M{\"u}ller, Klaus-Robert},
  journal={IEEE transactions on neural networks and learning systems},
  volume={28},
  number={11},
  pages={2660--2673},
  year={2016},
  publisher={IEEE}
}

@inproceedings{jing2024faithscore,
  title={Faithscore: Fine-grained evaluations of hallucinations in large vision-language models},
  author={Jing, Liqiang and Li, Ruosen and Chen, Yunmo and Du, Xinya},
  booktitle={Findings of the Association for Computational Linguistics: EMNLP 2024},
  pages={5042--5063},
  year={2024}
}

@inproceedings{es2024ragas,
  title={Ragas: Automated evaluation of retrieval augmented generation},
  author={Es, Shahul and James, Jithin and Anke, Luis Espinosa and Schockaert, Steven},
  booktitle={Proceedings of the 18th Conference of the European Chapter of the Association for Computational Linguistics: System Demonstrations},
  pages={150--158},
  year={2024}
}

@article{zhao2025grad,
  title={Grad-ECLIP: Gradient-based Visual and Textual Explanations for CLIP},
  author={Zhao, Chenyang and Wang, Kun and Hsiao, Janet H and Chan, Antoni B},
  journal={arXiv preprint arXiv:2502.18816},
  year={2025}
}

\appendix

\section{Implementation Details}
\label{app:implementation}

We use Llama-3.1-8B-Instruct and Mistral-7B-Instruct-v0.3 from Hugging Face Transformers \citep{wolf2020transformers}. A larger Llama-3.1-70B-Instruct evaluation is reported in the Appendix~\ref{app:larger_model}. Baselines are implemented with Inseq \citep{sarti-etal-2023-inseq}. We set \texttt{top\_p=1} for deterministic generation across attribution methods. Experiments with 7B/8B models are run on a single 48GB NVIDIA RTX6000 Ada GPU; the 70B model is evaluated on a single 80GB A100 with 4-bit quantization. Results are averaged over 15 Monte Carlo samples.


We generated heatmaps for all compared attribution methods using the Inseq library. Given an input prompt, we first obtained the model prediction and then computed token-level attribution scores for the input tokens with respect to the generated output. All Inseq-based methods were used with their default hyperparameters, except for Integrated Gradients and ReAGent. For Integrated Gradients, we used 16 integration steps with an internal batch size of 8, which provides a practical trade-off between attribution stability and GPU memory consumption. Since ReAGent relies on iterative probing and is significantly more expensive than gradient- or attention-based methods, we adopted a computationally feasible setting with \texttt{keep\_top\_n=3}, \texttt{stopping\_condition\_top\_k=3}, \texttt{replacing\_ratio=0.25}, \texttt{max\_probe\_steps=32}, and \texttt{num\_probes=2}. For each sample, the generated attribution scores were aggregated to obtain the final heatmap used in our evaluation.

\section{Derivation of Grad-ELLM}
\label{app:grad-ellm-derivation}

We provide the derivation of Grad-ELLM used in Sec.~\ref{sec:grad-ellm}. We focus on explaining the contribution of input tokens to $l_t$, the logit of the next generated token $y_t$. 
Analogous to Grad-ECLIP, we decompose the decoder to relate $l_t$ to intermediate features (see Fig. \ref{framework}). We first consider the last layer ($k=0$) where the final hidden state for the last token is $\bz^{(0)}_{t}$. The logit $l_t = \mathcal{LP}(\bz^{(0)}_{t})$, where $\mathcal{LP}$ is the linear projection to vocabulary space. Approximating linearity,
\begin{align}
       l_t &= \mathcal{LP}(\bz^{(0)}_{t})=\mathcal{LP}(\bo^{(0)}_{t} + \bz^{(1)}_{t}) \\
       &\approx \mathcal{LP}(\bo^{(0)}_{t})  + \mathcal{LP}(\bz^{(1)}_{t}) ,    
\end{align}
where $\bo^{(0)}_{t}$ is the $t$-th token in $\bo^{(0)}$, and
\begin{align}
    \bo^{(0)}_{t} &= \mathcal{A}(\bz^{(1)})_{t}
    = \sum_i \operatorname{softmax}( \tfrac{\mathbf{q}_t \cdot \mathbf{k}_i^\top}{\sqrt{d}}) \mathbf{v}_i,   
\end{align}
and $\mathcal{A}$ represents the self-attention layer. Thus we obtain the approximation of logit $l_t$ recursively,
\begin{align}
\label{linear approximation}
\nonumber
l_t &\approx  \mathcal{LP}(\bo^{(0)}_{t}) + \mathcal{LP}(\bo^{(1)}_{t}) + \cdots \\
   &\quad + \mathcal{LP}(\bo^{(N-1)}_{t})  
     + \mathcal{LP}(\bz^{(N)}_{t}) 
    \overset{\triangle}{=} \sum_k l_t^{(k)}  ,
\end{align}
as an aggregation of features from each layer.

Following \citet{zhao2025grad}, and looking at the last transformer layer as an example, we approximate the logit of the next predicted token as a weighted combination of the channel features in $\bo_{t}$. 
For the last layer ($k=0$),
\begin{align}
    l_t^{(0)} &= \mathcal{LP}(\bo_t^{(0)}) = \sum_c \mathcal{LP}(\bo_{t})[c]^{(0)} 
     \overset{\triangle}{=} f(\bo_t),
\end{align}
where we have defined the logit of the next generated token as a function of $\bo_t$, i.e., $f(\bo_t)$. We denote the linear approximation of the logit as $\tilde{f}$, 
\begin{align}
    f(\bo_t) \approx \tilde{f}(\bo_t) \overset{\triangle}{=} \sum_c w_c o_t[c] = \bw \bo_t^{\top}
\end{align}
The linear weights $\bw$ are obtained by matching the first derivatives of $f$ and its linear approximation:
\begin{align}
        \bw &= \underset{w}{\operatorname{argmin}} \Vert f'(\bo_{t}) - \tilde{f}'(\bo_{t}) \Vert^2 \\
          &= \operatorname{argmin} \Vert \tfrac{\partial{f}}{\partial{\bo_{t}}} - \bw \Vert^2 ,
\end{align}
and thus we obtain the solution $\bw^* = \tfrac{\partial{f}}{\partial{\bo_{t}}}$.
Finally, combining with (\ref{spatial weight}) we have the approximation:

\begin{align}
        f(\bo_{t}) &\approx \sum_c w_c o_{t} [c] \\
        &= \sum_c \tfrac{\partial{f(\bo_{t})}}{\partial{\bo_{t}}} \sum_i \operatorname{softmax}(\tfrac{\mathbf{q}_{t} \cdot \mathbf{k}_i^\top}{\sqrt{d}})\mathbf{v}_{ic} \\
        &= \sum_i \Big[\sum_c \underbrace{\tfrac{\partial{f(\bo_{t})}}{\partial{\bo_{t}}}}_{w_c} \underbrace{\operatorname{softmax} (\tfrac{\mathbf{q}_{t} \cdot \mathbf{k}_i^\top}{\sqrt{d}})}_{\lambda_i}\mathbf{v}_{ic} \Big],
\end{align}
where $\mathbf{v}_{ic}$ is the $c$-th channel of $\mathbf{v}_i$. Thus the attribution for the $i$-th token is
\begin{equation}
    H_i = \operatorname{ReLU}(\sum_c w_c \lambda_i \mathbf{v}_{ic}),
\end{equation}
where $\operatorname{ReLU}$ is used to focus only on tokens with positive influence on the logit value. In this way we decompose the attribution into two parts: channel weights $w_c$ and token weights $\lambda_i$. Following \citet{zhao2024gradient}, the attribution map will be obtained by $\bH = [H_i]_i$ using the last layer value $v$ as the feature map.  Based on (\ref{linear approximation}), we can select the desired number of layers to aggregate information from different layers to obtain the heatmap.

As with \citet{zhao2024gradient}, we also apply 0-1 normalization on the similarities $[\mathbf{q}_t\mathbf{k}_i^\top]_i$ and use it to replace the original $\operatorname{softmax}$, i.e., $\lambda_i \approx \Phi(\mathbf{q}_t\mathbf{k}_i^\top)$, where $\Phi$ is the 0-1 normalization function applied over the set of similarities. Without this loosening step, the softmax operation will produce sparser heatmaps, due to the peakiness of softmax. The loosening step helps to reveal unattended input tokens that are similar to the attended token, which likely contain similar information. This design makes Grad-ELLM different from attribution methods that aim to produce extremely sparse token rankings. By combining attention-derived token weights with gradient-derived channel weights, Grad-ELLM produces a continuous attribution distribution over the input.

\section{Dataset Details}
\label{app:dataset}

More detailed descriptions of the datasets:

\begin{compactitem}
    \item \textbf{IMDb} is introduced by \cite{maas2011learning} which is a widely used benchmark for binary sentiment classification. It consists of 50,000 movie reviews collected from the Internet Movie Database (IMDb), each labeled as either positive or negative. We randomly sampled 1000 positive examples and 1000 negative examples from them.
    \item \textbf{SST2} is also a sentiment classification task introduced by \citet{socher2013recursive}, which consists of independent single sentences, many of which contain complex grammatical structures, negation, and irony, making classification more challenging compared with IMDb. We also randomly sampled 1000 entries from each of the positive and negative samples.
    \item \textbf{BoolQ} is a dataset from the ERASER \cite{deyoung-etal-2020-eraser} benchmark, which tests whether a model can read a paragraph and correctly answer a factual yes/no question about it, emphasizing deep language understanding over superficial cues.
    \item \textbf{TellMeWhy} is a generation dataset for answering questions in narratives. We follow the dataset setting in \cite{zhao2024reagent, lal-etal-2021-tellmewhy}.
    \item \textbf{WikiBio} is a dataset consisting of Wikipedia biographies. Following \citet{zhao2024reagent, manakul2023selfcheckgpt}, we take the first two sentences as a prompt.
\end{compactitem}

\section{Prompt Details}
\label{sec:appendix}

\begin{table*}[htbp]
\tiny
\centering
\begin{tabular}{cp{0.8\textwidth}}
\toprule
\centering 
\textbf{Dataset} & \textbf{Prompt Construction Code} \\
\hline
\\
\adjustbox{}{IMDb \& SST2} & 
\adjustbox{}{%
\begin{minipage}{\linewidth}
\centering 
\begin{verbatim}
messages = [
    {
        "role": "system",
        "content": "You are a helpful sentiment classifier. Please help to do the sentiment classification
        of the given text and respond ONLY with the single word Positive or Negative.",
    },
    {"role": "user", "content": f"Text: {text}"},
            ]
\end{verbatim}
\end{minipage}} \\
\\
\hline
\\
\adjustbox{}{BoolQ} & 
\adjustbox{}{%
\centering 
\begin{minipage}{\linewidth}
\begin{verbatim}
messages = [
    {
        "role": "system",
        "content": "You are a factual question answering system. "
                "Determine if the given passage supports the question being true. "
                "Respond ONLY with the single word 'true' or 'false' in lowercase, "
                "without any punctuation or additional text."
    },
    {
        "role": "user",
        "content": "Passage: {passage}\n\nQuestion: {question}".format(
            passage=passage_text,
            question=question_text
        )
    }
            ]
\end{verbatim}
\end{minipage}} \\
\\
\hline
\\
\adjustbox{}{TellMeWhy \& WikiBio} & 
\adjustbox{}{%
\centering 
\begin{minipage}{\linewidth}
\begin{verbatim}
messages = [
    {
        "role": "user",
        "content": text
    }
            ]
\end{verbatim}
\end{minipage}} \\
\bottomrule
\\
\end{tabular}
\caption{Detailed prompts for all datasets}
\label{tab:prompt-final}
\end{table*}

The full prompts are shown in Tab. \ref{tab:prompt-final}. For Llama, we add some special tokens at the end of the input text to ensure that the model outputs text in the desired format. These special tokens are: "<|start\_header\_id|> assistant <|end\_header\_id|> \textbackslash n".
These tokens correspond to the assistant header expected by the Llama chat template. If we don't force the model to output them, it might not output them according to our format, which could lead to incorrect output during deletion and insertion evaluation. However, this requirement does not apply to Mistral.

\section{Evaluation with Insertion/Deletion}
\label{app:insertion/deletion}

For completeness, we also compute the classical Deletion and Insertion \cite{DBLP:conf/bmvc/PetsiukDS18} metrics for classification tasks.
%
Inspired by \citet{deyoung-etal-2020-eraser}, to more robustly evaluate the performance of attribution methods, we evaluate the change in flip probability rather than the change in prediction probability. In our experiment setting, we set each perturbation step to 5\% of the tokens and record results for 20 steps.

Higher AUC for Deletion is better, since ideally removing words will yield large flip probability (large performance degradation).
Conversely, lower AUC for Insertion is better, since ideally adding words will yield a reduction in flip probability (large performance gain).
%
It is worth noting that if the change in prediction probability is the evaluation target, Value Zeroing should theoretically be the upper bound for Deletion and insertion experiments, as it directly measures the causal effect of zeroing features on the model's prediction. However, within the evaluation framework of flip probability, this upper bound property may be weakened, because flip probability focuses on crossing class boundaries rather than the absolute change in probability magnitude.

\begin{table*}
  \centering
  \tiny
  \begin{tabular}{cccccccccc}
    \hline
    \textbf{Deletion$\uparrow$}&Attention&DeepLIFT&Input$\times$Gradients&Integrated Gradients&
    Random&Saliency&Value Zeroing&ReAGent&Ours (w/ | w/o)\\
    \hline
    IMDb& 0.2572 & 0.1956 & 0.1929 & 0.1690 & 0.1772 & 0.1681 & 0.2618 & 0.1820 & 0.2153 | 0.2520 \\
    SST2& 0.3310 & 0.3050 & 0.3058 & 0.2821 & 0.2868 & 0.2741 & 0.3462 & 0.2987 & 0.3251 | 0.3383 \\
    BoolQ& 0.2393 & 0.2958 & 0.2956 & 0.2852 & 0.2292 & 0.2955 & 0.2490 & 0.2299 & 0.2673 | 0.2621 \\
    Average& \textcolor{green}{0.2758} & 0.2655 & 0.2648 & 0.2454 & 0.2311 & 0.2459 & \textcolor{red}{0.2857} & 0.2369 & 0.2692 | \textcolor{blue}{0.2841}  \\
    \hline
    \textbf{Insertion$\downarrow$}&Attention&DeepLIFT&Input$\times$Gradients&Integrated Gradients&
    Random&Saliency&Value Zeroing&ReAGent&Ours (w/ | w/o)\\
    \hline
    IMDb& 0.0847 & 0.1242 & 0.1294 & 0.1275 & 0.1718 & 0.1358 & 0.0832 & 0.1706 & 0.1336 | 0.1021 \\
    SST2& 0.1816 & 0.2022 & 0.2022 & 0.2174 & 0.2579 & 0.2217 & 0.1754 & 0.2532 & 0.2160 | 0.1957 \\
    BoolQ& 0.1885 & 0.0991 & 0.1005 & 0.1137 & 0.2087 & 0.0995 & 0.1902 & 0.2106 & 0.1740 | 0.1764 \\
    Average& 0.1516 & \textcolor{red}{0.1418} & \textcolor{blue}{0.1440} & 0.1529 & 0.2128 & 0.1523 & \textcolor{green}{0.1496} & 0.2115 & 0.1745 | 0.1581  \\
    \hline
  \end{tabular}
  \caption{\label{llama-deletion-insertion}
    Faithfulness evaluation of Deletion and Insertion on IMDb, SST2, and BoolQ with Llama. The best performance, second place, and third place are marked in \textcolor{red}{red}, \textcolor{blue}{blue}, and \textcolor{green}{green} respectively. w/ denotes loosened attention, and w/o denotes original attention.
  }
\end{table*}

\begin{table*}
  \centering
  \tiny
  \begin{tabular}{cccccccccc}
    \hline
    \textbf{Deletion$\uparrow$}&Attention&DeepLIFT&Input$\times$Gradients&Integrated Gradients&
    Random&Saliency&Value Zeroing&ReAGent&Ours (w/ | w/o)\\
    \hline
    IMDb& 0.2719 & 0.1967 & 0.1951 & 0.1726 & 0.1892 & 0.1898 & 0.2415 & 0.1795 & 0.2296 | 0.2266 \\
    SST2& 0.3550 & 0.2821 & 0.2839 & 0.3013 & 0.2903 & 0.2828 & 0.3486 & 0.2857 & 0.3167 | 0.3226 \\
    BoolQ& 0.2776 & 0.3042 & 0.3023 & 0.2812 & 0.2339 & 0.3055 & 0.2579 & 0.2339 & 0.2658 | 0.2606 \\
    Average& \textcolor{red}{0.3015} & 0.2610 & 0.2604 & 0.2517 & 0.2378 & 0.2593 & \textcolor{blue}{0.2826} & 0.2330 & \textcolor{green}{0.2707} | 0.2699  \\
    \hline
    \textbf{Insertion$\downarrow$}&Attention&DeepLIFT&Input$\times$Gradients&Integrated Gradients&
    Random&Saliency&Value Zeroing&ReAGent&Ours (w/ | w/o)\\
    \hline
    IMDb& 0.0861 & 0.1079 & 0.1074 & 0.1189 & 0.1849 & 0.1062 & 0.0973 & 0.1806 & 0.1200 | 0.1273 \\
    SST2& 0.1863 & 0.2127 & 0.2121 & 0.2073 & 0.2652 & 0.2089 & 0.1899 & 0.2532 & 0.2023 | 0.2047 \\
    BoolQ& 0.1556 & 0.1067 & 0.1079 & 0.1310 & 0.2226 & 0.1044 & 0.1974 & 0.2191 & 0.1983 | 0.1953 \\
    Average& 0.1427 & \textcolor{blue}{0.1424} & \textcolor{green}{0.1425} & 0.1524 & 0.2242 & \textcolor{red}{0.1398} & 0.1615 & 0.2177 & 0.1735 | 0.1757  \\
    \hline
  \end{tabular}
  \caption{\label{mistral-deletion-insertion}
    Faithfulness evaluation of Deletion and Insertion on IMDb, SST2, and BoolQ with Mistral. The best performance, second place, and third place are marked in \textcolor{red}{red}, \textcolor{blue}{blue}, and \textcolor{green}{green} respectively. w/ denotes loosened attention, and w/o denotes original attention.
  }
\end{table*}

\subsection{Results}

The Insertion/Deletion AUC results are presented in Tabs.\ref{llama-deletion-insertion} and  \ref{mistral-deletion-insertion}. 
%
In the deletion experiments, our method achieved the second-best result with Llama and the third-best result with Mistral. However, our method yielded slightly worse results in the insertion experiments. Through in-depth analysis, we found that our heatmap is denser compared to other methods, which puts us at a disadvantage in order-based evaluations such as deletion and insertion, because we are more susceptible to small noise that swaps the order of words. 
On the other hand, as presented in the main paper, our $\pi$-Soft-NC/NS metrics reveal that our method has better overall importance distributions.

\section{Additional Qualitative Results}
\label{app:qualitative_full_examples}

We provide additional qualitative examples to complement Fig.~\ref{fig:qualitative_sentiment_examples}. 
Since IMDb reviews can be substantially longer than SST2 sentences, the main paper only shows the key sentence from the negative IMDb example. 
Here, we show the full IMDb review corresponding to Fig.~\ref{fig:qualitative_sentiment_examples}(b), as well as another negative IMDb example with a similar attribution pattern.

\begin{figure*}[h]
  \centering
  \includegraphics[width=0.98\textwidth]{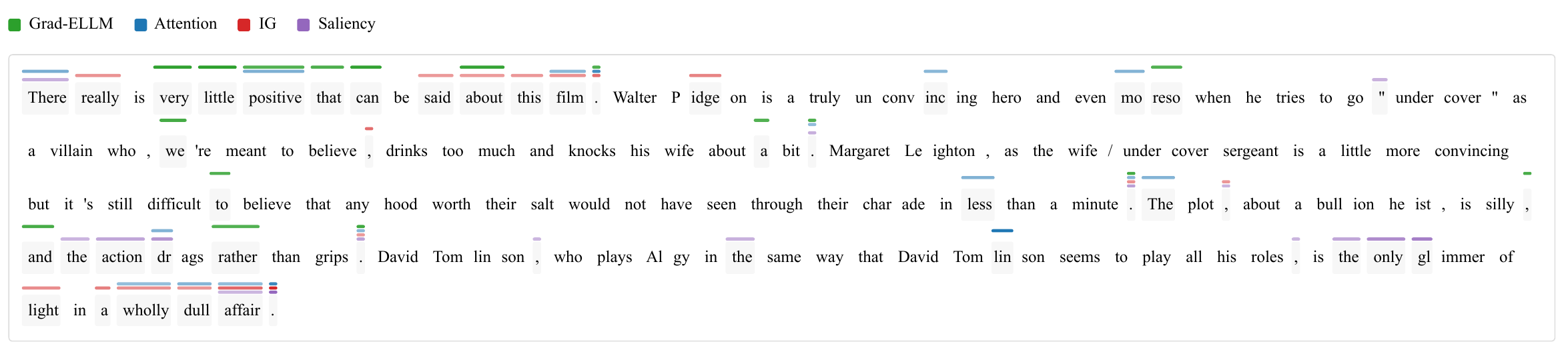}
  \caption{
  Full review for the negative IMDb example shown in Fig.~\ref{fig:qualitative_sentiment_examples}(b).
  Colored bars indicate tokens selected among the top 10\% attribution scores by each method.
  The full context shows that the key phrase ``very little positive'' is part of a longer negative review.
  }
  \label{fig:imdb_negative_full_example}
\end{figure*}

Fig.~\ref{fig:imdb_negative_full_example} shows the full review for the negative IMDb example in the main paper. 
The key phrase ``very little positive'' appears in the full context, and Grad-ELLM still selects the phrase-level evidence rather than only the isolated token ``positive''. 
This suggests that Grad-ELLM can capture compositional sentiment cues within a longer review.

\begin{figure*}[h]
  \centering
  \includegraphics[width=0.98\textwidth]{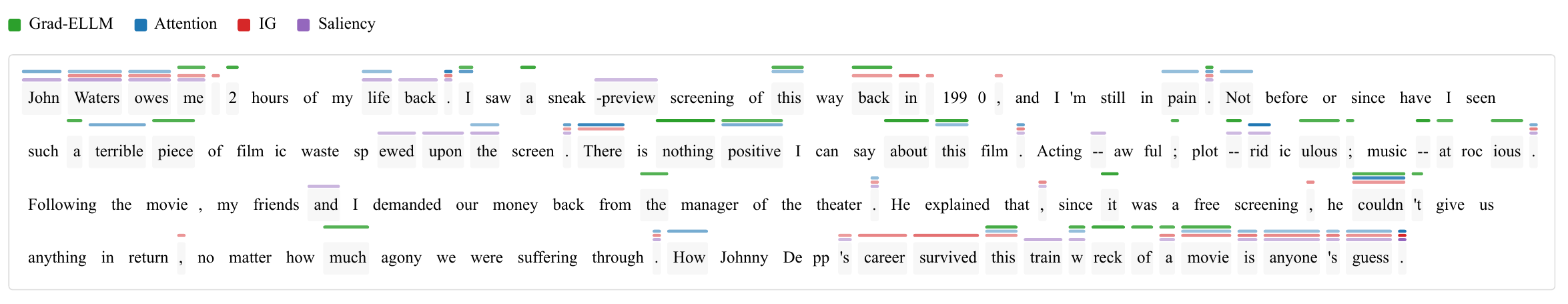}
  \caption{
  Additional negative IMDb example. Colored bars indicate tokens selected among the top 20\% attribution scores by each method.
  }
  \label{fig:imdb_negative_additional_example}
\end{figure*}

Fig.~\ref{fig:imdb_negative_additional_example} shows another negative IMDb example. 
In this review, Grad-ELLM highlights multiple strongly negative phrases, such as ``nothing positive'', ``terrible piece of film'', and related negative expressions. 
This example further illustrates that the method does not merely select sentiment words in isolation, but can highlight broader phrases that support the model's negative prediction.

\section{Results on Larger Models}
\label{app:larger_model}

We also conduct experiments with Llama-3.1-70B-Instruct to evaluate whether the proposed metrics and Grad-ELLM remain effective on a larger model. The 
results are shown in Tab.~\ref{llama70b-deletion/insertion}. Overall, the 70B results show a pattern consistent with the main 7B/8B experiments: Grad-ELLM is especially strong under the comprehensiveness-oriented $\pi$-Soft-NC metric, achieving the best average $\pi$-Soft-NC AUC across datasets. This suggests that Grad-ELLM's attribution distribution continues to provide broad coverage of important evidence when scaling to a larger LLM.

On $\pi$-Soft-NS, Grad-ELLM remains competitive but is not the top-performing method on average. This again indicates that sufficiency-oriented evaluation and comprehensiveness-oriented evaluation capture different attribution behaviors. Sparse methods such as Attention, Integrated Gradients, or Saliency can sometimes better preserve the prediction with a smaller subset of retained tokens, while Grad-ELLM is more effective at identifying evidence whose removal causes a larger distributional shift. These results further support our main argument that faithful attribution should be evaluated along multiple axes rather than by a single metric.

\section{Computational Complexity}
We analyze the average attribution time per single sample for different methods on the IMDb dataset. The results are shown in Tab.~\ref{computational_complexity}. As shown in the table, our gradient-based method is faster than most of the compared methods while achieving competitive results. Note that ReAGent and Value Zeroing are much slower because they require multiple token replacements and forward passes. More specifically, Value Zeroing requires over 45.2 seconds per sample, while our method only takes 0.32 seconds.

\begin{table}[htbp]
\centering
\small
\begin{tabular}{lrr}
\toprule
Method & Llama-8B & Mistral-7B \\
\midrule
Attention & \textbf{0.31} & \textbf{0.32} \\
DeepLIFT & 0.95 &  1.35 \\
Input$\times$Gradient & 0.63 & 0.80 \\
Integrated Gradients & 6.45 & 9.05 \\
Saliency & 0.63 & 0.81 \\
Value Zeroing & 45.20 & 47.80 \\
ReAGent & 17.50 & 18.70 \\
Ours & 0.32 & 0.38 \\
\bottomrule
\end{tabular}
\caption{Average Time per Sample (Seconds) for generating heatmaps on IMDb (Llama-3.1-8B-Instruct and Mistral-7B-Instruct-v0.3). Times averaged over evaluation set.}
\label{computational_complexity}
\end{table}

\begin{table*}
  \centering
  \tiny
  \begin{tabular}{ccccccccc}
    \hline
    \textbf{$\pi$-Soft-NS$\uparrow$}
    & Attention
    & DeepLIFT
    & Input$\times$Gradients
    & Integrated Gradients
    & Random
    & Saliency
    & Value Zeroing
    & Ours\\
    \hline
    IMDb      & 0.7348 & 0.6937 & 0.6982 & 0.7447 & 0.7069 & 0.7036 & 0.7205 & 0.7152\\
    SST2      & 0.6763 & 0.6230 & 0.6296 & 0.6839 & 0.5949 & 0.6299 & 0.6432 & 0.6586\\
    BoolQ     & 0.5090 & 0.5235 & 0.5232 & 0.5210 & 0.4273 & 0.5374 & 0.5119 & 0.4834\\
    TellMeWhy & 0.6369 & 0.6277 & 0.6320 & 0.5723 & 0.4513 & 0.6415 & 0.6004 & 0.6152\\
    WikiBio   & 0.5369 & 0.4808 & 0.4899 & 0.4812 & 0.3687 & 0.5009 & 0.4872 & 0.4934\\
    Average   & \textcolor{red}{0.6187} & 0.5897 & 0.5945 & \textcolor{green}{0.6006} & 0.5098 & \textcolor{blue}{0.6026} & 0.5926 & 0.5931\\
    \hline

    \textbf{$\pi$-Soft-NC$\uparrow$}
    & Attention
    & DeepLIFT
    & Input$\times$Gradients
    & Integrated Gradients
    & Random
    & Saliency
    & Value Zeroing
    & Ours\\
    \hline
    IMDb      & 0.2112 & 0.1607 & 0.1699 & 0.1849 & 0.1822 & 0.1654 & 0.1880 & 0.2068\\
    SST2      & 0.3337 & 0.2619 & 0.2662 & 0.3124 & 0.3082 & 0.2803 & 0.2746 & 0.3228\\
    BoolQ     & 0.4718 & 0.4622 & 0.4640 & 0.4819 & 0.4962 & 0.4892 & 0.4820 & 0.5343\\
    TellMeWhy & 0.4769 & 0.3985 & 0.4004 & 0.3510 & 0.3829 & 0.4203 & 0.3752 & 0.4305\\
    WikiBio   & 0.5853 & 0.5685 & 0.5617 & 0.4976 & 0.5135 & 0.5708 & 0.5694 & 0.6060\\
    Average   & \textcolor{blue}{0.4157} & 0.3703 & 0.3724 & 0.3655 & 0.3766 & \textcolor{green}{0.3852} & 0.3778 & \textcolor{red}{0.4200}\\
    \hline
  \end{tabular}
  \caption{\label{llama70b-deletion/insertion}
  Faithfulness evaluation with AUC of $\pi$-Soft-NC/NS curves for Llama-3.1-70B-Instruct. The best performance, second place, and third place are marked in \textcolor{red}{red}, \textcolor{blue}{blue}, and \textcolor{green}{green} respectively. 
  }
\end{table*}

\section{Layer Aggregation Analysis}
\label{app:layer-ablation}

Grad-ELLM aggregates attribution information from multiple decoder layers, as described in Eq.~(5).
The number of aggregated layers controls how much intermediate attribution evidence is incorporated into the final heatmap.
Since attribution behavior can vary across datasets and model families, we do not fix this number to all decoder layers.
Instead, we perform a lightweight search over a small candidate set of layer numbers and select the configuration that performs best for each dataset and model.

Specifically, we evaluate Grad-ELLM with the number of aggregated layers selected from
\(\{1, 4, 8, 16, 32\}\).
For each setting, we compute the AUC of the \(\pi\)-Soft-NS and \(\pi\)-Soft-NC curves using the same evaluation protocol as in the main experiments.
The purpose of this analysis is to examine how sensitive Grad-ELLM is to the layer aggregation depth and to justify the layer configurations used in the main results.

\begin{figure*}[t]
    \centering
    \includegraphics[width=0.48\textwidth]{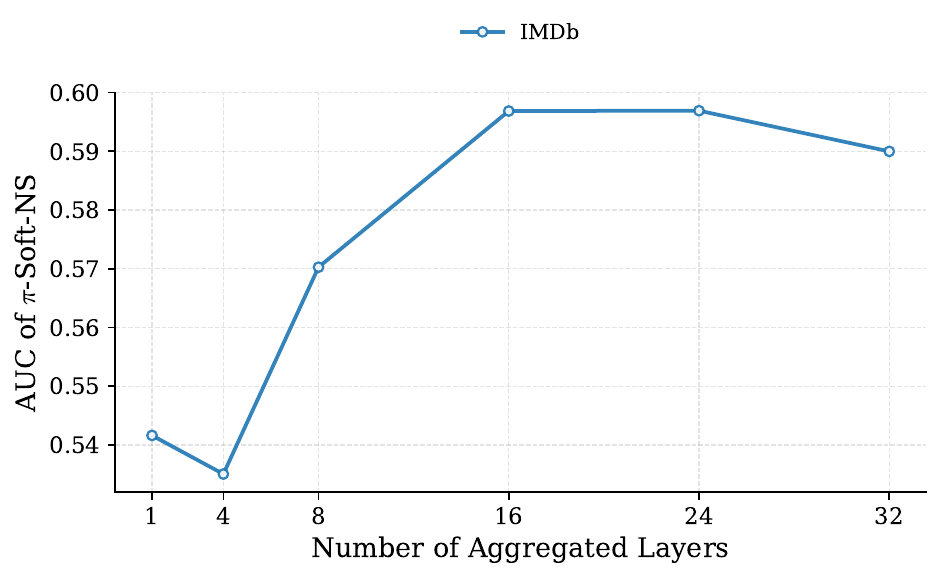}
    \includegraphics[width=0.48\textwidth]{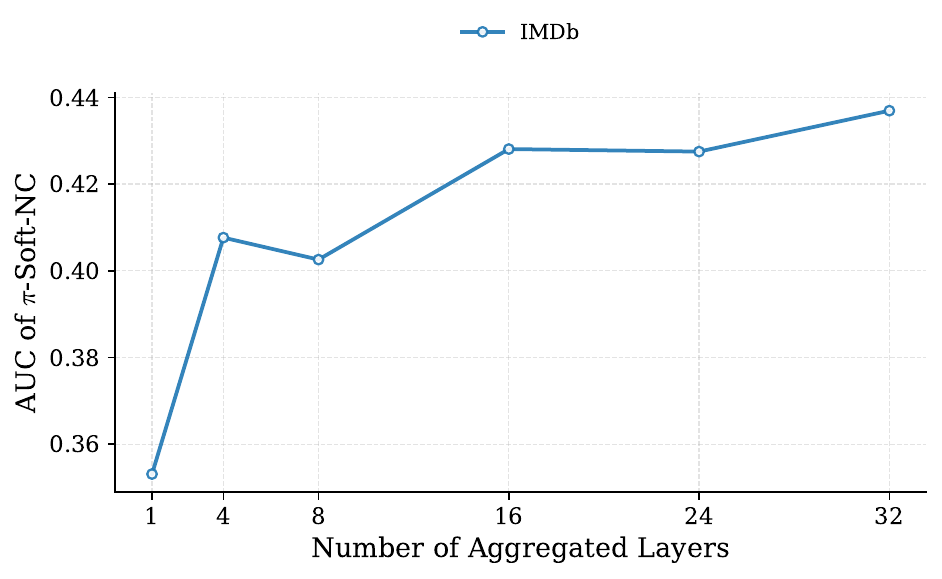}
    \caption{
    Layer aggregation analysis for Grad-ELLM on IMDb with Llama.
    The x-axis denotes the number of aggregated decoder layers, and the y-axis denotes the AUC of the corresponding \(\pi\)-Soft metric.
    Left: AUC of \(\pi\)-Soft-NS.
    Right: AUC of \(\pi\)-Soft-NC.
    }
    \label{fig:layer-ablation}
\end{figure*}

As shown in Fig.~\ref{fig:layer-ablation}, the optimal number of aggregated layers is not always the maximum number of layers.
This indicates that attribution evidence is distributed differently across layers depending on the dataset and metric.
Using too few layers may miss useful intermediate evidence, while aggregating too many layers can also introduce noisy or less task-relevant attribution signals.
Therefore, in the main experiments, we select the layer aggregation depth from a small candidate set rather than always using all layers.

\begin{figure}[htb]
\centering
\includegraphics[width=0.92\linewidth]{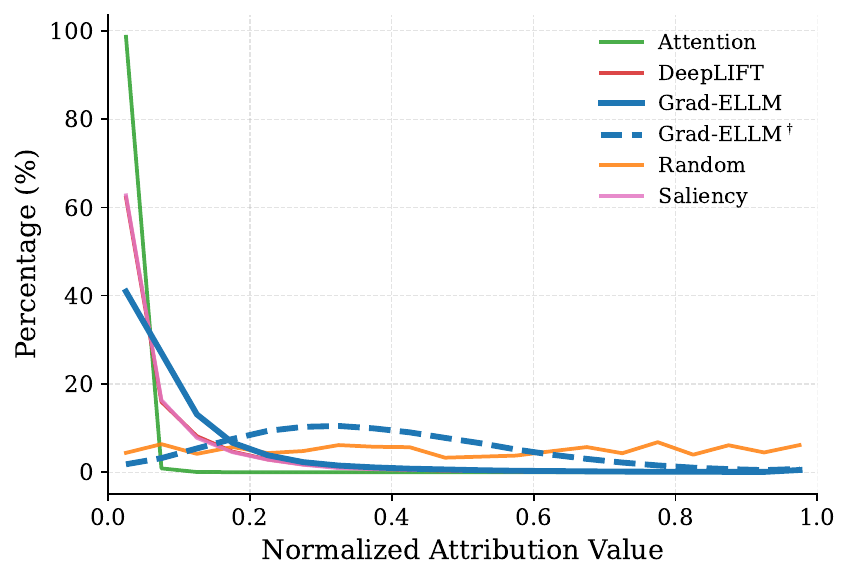}
\caption{
Attribution-value distributions across different methods with Llama on IMDb. Grad-ELLM$^\dagger$ denotes the variant with loosened attention normalization. Random is dense and nearly uniform but performs poorly under $\pi$-Soft metrics, indicating that density alone does not guarantee high faithfulness.
}
\label{fig:attribution-distribution}
\end{figure}

\section{Attribution Distribution Analysis}
\label{app:attribution-distribution}

To examine whether $\pi$-Soft-NC/NS simply favor dense attribution maps, we visualize the distribution of attribution values across different methods. Fig.~\ref{fig:attribution-distribution} shows that Grad-ELLM produces a more evenly distributed attribution map than sparse gradient-based methods such as DeepLIFT and Saliency. However, density alone does not explain faithfulness performance: Random produces the most uniform attribution distribution but performs poorly under $\pi$-Soft metrics.

This suggests that the proposed metrics do not merely reward dense heatmaps. Instead, strong $\pi$-Soft performance requires calibrated selectivity, where attribution scores distribute probability mass broadly enough to cover relevant evidence while still distinguishing informative tokens from uninformative ones.

\section{Fixed-Routing Causal Evaluation}
\label{app:fixed-routing}


Here we present the results using the \emph{fixed-routing} protocol. 
In particular, for each input, we first run the model on the original unperturbed sequence and cache the attention maps. During the evaluation of softly perturbed inputs, we reuse the cached attention maps, so that the attention routing remains the same as in the original forward pass. Under this protocol, changes in the output distribution are mainly caused by changes to the value/content information carried by the perturbed tokens, rather than the potential influence of attention reallocation.

\begin{figure*}[htb]
  \centering
  \vspace{-0.8em}

  \begin{minipage}[t]{0.485\textwidth}
    \centering
    \includegraphics[width=\linewidth]{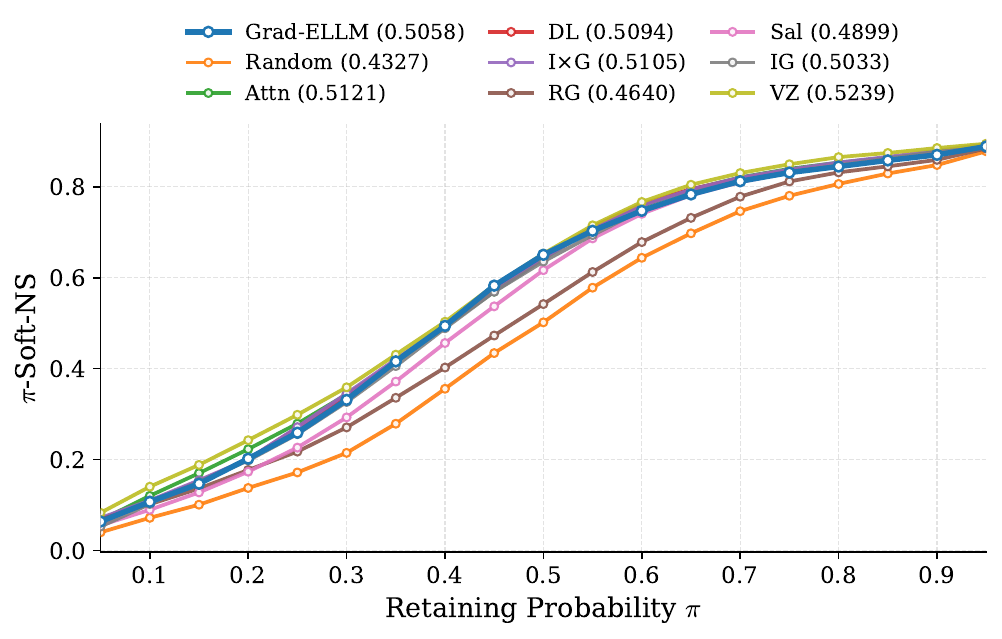}\\[-0.8em]
    {\small (a) $\pi$-Soft-NS vs. Retaining Probability}
  \end{minipage}
  \hfill
  \begin{minipage}[t]{0.485\textwidth}
    \centering
    \includegraphics[width=\linewidth]{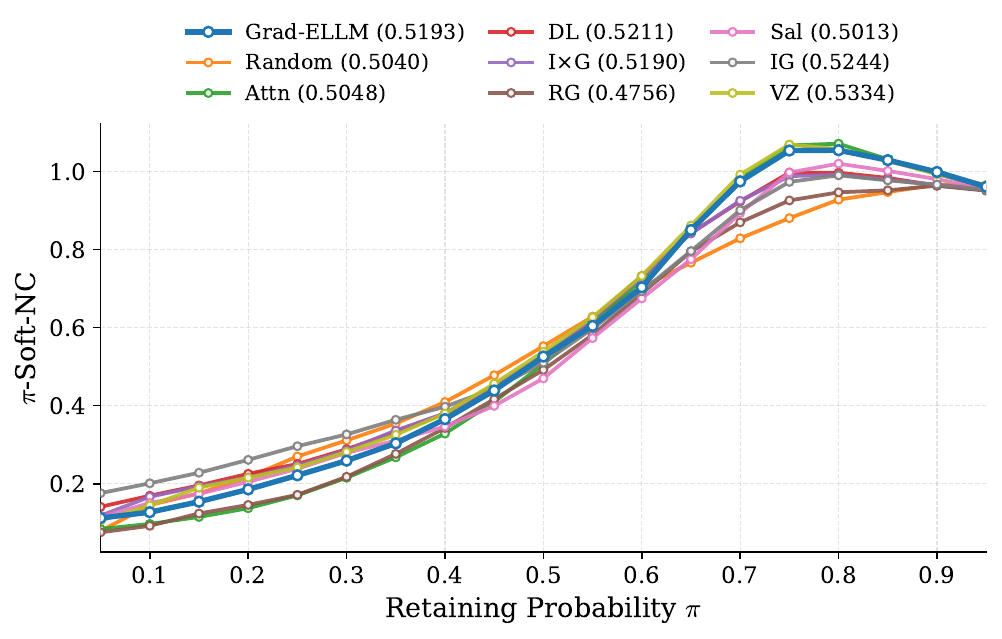}\\[-0.8em]
    {\small (b) $\pi$-Soft-NC vs. Retaining Probability}
  \end{minipage}

  \vspace{-0.6em}
  \caption{Fixed-Routing version of proposed $\pi$-Soft-NC/NS vs. Retaining Probability on SST2 for different XAI methods with Llama.}
  \label{soft-ns/nc_fig_freezing}
  \vspace{-1.0em}
\end{figure*}

The results are presented in Table \ref{tab:fixed-routing} and Fig.~\ref{soft-ns/nc_fig_freezing}. Input$\times$Gradient, Saliency and DeepLift perform well under the $\pi$-Soft-NS metric. This is likely due to their use of input-level gradients that measure the effect of an input token on the output.
On the other hand, there is no clearly dominant method for $\pi$-Soft-NC under fixed-routing. 
Note that several methods consistently outperform raw attention under the fixed-routing protocol. This indicates that attention itself is lacking as an attribution method -- some attended words do not have large affects on the output. 

\begin{table*}[t]
\centering
\tiny
\setlength{\tabcolsep}{3.0pt}
\renewcommand{\arraystretch}{1.05}

\begin{subtable}[t]{0.49\textwidth}
\centering
\caption{Llama: AUC of $\pi$-Soft-NC/NS under fixed-routing evaluation ($\uparrow$)}
\label{tab:llama-fixed-routing}
\begin{tabular}{lccccccccc}
\toprule
 & \multicolumn{9}{c}{Methods} \\
\cmidrule(lr){2-10}
Dataset & {Attn} & {DL} & {I$\times$G} & {IG} & {Rnd} & {Sal} & {VZ} & {RG} & {Ours} \\
\midrule
\multicolumn{10}{l}{\textbf{AUC $\pi$-Soft-NS $\uparrow$}}\\
IMDb      & 0.649 & 0.633 & 0.632 & 0.629 & 0.582 & 0.624 & 0.645 & 0.596 & 0.644 \\
SST2      & 0.512 & 0.509 & 0.510 & 0.503 & 0.433 & 0.490 & 0.524 & 0.464 & 0.517 \\
BoolQ     & 0.386 & 0.455 & 0.452 & 0.425 & 0.366 & 0.450 & 0.389 & 0.364 & 0.412 \\
TellMeWhy & 0.644 & 0.668 & 0.665 & 0.647 & 0.520 & 0.663 & 0.649 & 0.555 & 0.625 \\
WikiBio   & 0.653 & 0.654 & 0.652 & 0.649 & 0.506 & 0.648 & 0.663 & 0.550 & 0.624 \\
\textbf{Avg} & 0.569 & \best{0.584} & \second{0.582} & 0.571 & 0.482 & \third{0.575} & 0.574 & 0.506 & 0.566 \\

\midrule
\multicolumn{10}{l}{\textbf{AUC $\pi$-Soft-NC $\uparrow$}}\\
IMDb      & 0.348 & 0.314 & 0.312 & 0.318 & 0.309 & 0.309 & 0.339 & 0.299 & 0.320 \\
SST2      & 0.505 & 0.521 & 0.519 & 0.524 & 0.504 & 0.501 & 0.533 & 0.476 & 0.519 \\
BoolQ     & 0.571 & 0.704 & 0.698 & 0.644 & 0.691 & 0.692 & 0.576 & 0.637 & 0.675 \\
TellMeWhy & 0.562 & 0.572 & 0.566 & 0.527 & 0.397 & 0.566 & 0.572 & 0.395 & 0.484 \\
WikiBio   & 0.593 & 0.521 & 0.519 & 0.543 & 0.428 & 0.511 & 0.638 & 0.411 & 0.502 \\
\textbf{Avg} & 0.516 & \second{0.526} & \third{0.523} & 0.511 & 0.466 & 0.516 & \best{0.532} & 0.443 & 0.500 \\
\bottomrule
\end{tabular}

\vspace{2pt}
\captionsetup{font=footnotesize}
\end{subtable}
\hfill
\begin{subtable}[t]{0.49\textwidth}
\centering
\caption{Mistral: AUC of $\pi$-Soft-NC/NS under fixed-routing evaluation ($\uparrow$)}
\label{tab:mistral-fixed-routing}
\begin{tabular}{lccccccccc}
\toprule
& \multicolumn{9}{c}{Methods} \\
\cmidrule(lr){2-10}
Dataset & {Attn} & {DL} & {I$\times$G} & {IG} & {Rnd} & {Sal} & {VZ} & {RG} & {Ours} \\
\midrule
\multicolumn{10}{l}{\textbf{AUC $\pi$-Soft-NS $\uparrow$}}\\
IMDb      & 0.612 & 0.627 & 0.626 & 0.626 & 0.584 & 0.631 & 0.579 & 0.588 & 0.622 \\
SST2      & 0.502 & 0.499 & 0.499 & 0.517 & 0.485 & 0.498 & 0.503 & 0.478 & 0.517 \\
BoolQ     & 0.455 & 0.482 & 0.481 & 0.469 & 0.433 & 0.484 & 0.445 & 0.426 & 0.446 \\
TellMeWhy & 0.511 & 0.520 & 0.521 & 0.485 & 0.400 & 0.522 & 0.512 & 0.423 & 0.469 \\
WikiBio   & 0.629 & 0.628 & 0.629 & 0.609 & 0.480 & 0.632 & 0.627 & 0.519 & 0.583 \\
\textbf{Avg} & 0.542 & \third{0.551} & \second{0.551} & 0.541 & 0.476 & \best{0.553} & 0.533 & 0.487 & 0.527 \\

\midrule
\multicolumn{10}{l}{\textbf{AUC $\pi$-Soft-NC $\uparrow$}}\\
IMDb      & 0.404 & 0.411 & 0.410 & 0.404 & 0.356 & 0.406 & 0.388 & 0.321 & 0.405 \\
SST2      & 0.711 & 0.728 & 0.730 & 0.779 & 1.031 & 0.891 & 0.719 & 1.262 & 1.076 \\
BoolQ     & 0.465 & 0.509 & 0.505 & 0.475 & 0.460 & 0.517 & 0.435 & 0.409 & 0.465 \\
TellMeWhy & 0.800 & 0.789 & 0.798 & 0.766 & 0.678 & 0.789 & 0.817 & 0.623 & 0.825 \\
WikiBio   & 0.667 & 0.611 & 0.611 & 0.582 & 0.485 & 0.616 & 0.686 & 0.494 & 0.547 \\
\textbf{Avg} & 0.609 & 0.609 & 0.611 & 0.601 & 0.602 & \second{0.644} & 0.609 & \third{0.622} & \best{0.664} \\
\bottomrule
\end{tabular}
\end{subtable}

\caption{Fixed-routing protocol with AUC of $\pi$-Soft-NC/NS curves for (a) Llama and (b) Mistral. Attention maps are cached from the original input and reused when evaluating softly perturbed inputs. This protocol tests whether the token content selected by each attribution method remains causally important under the original attention routing. \textcolor{red}{Best} / \textcolor{blue}{2nd} / \textcolor{green}{3rd} are highlighted on the \textbf{Avg} row. Abbr.: Attn = Attention, DL = DeepLIFT, I$\times$G = Input$\times$Gradients, IG = Integrated Gradients, Rnd = Random, Sal = Saliency, VZ = Value Zeroing, RG = ReAGent.}
\label{tab:fixed-routing}
\end{table*}

\end{document}